\documentclass[runningheads]{llncs}

\usepackage{eccv}

\usepackage{eccvabbrv}

\usepackage{graphicx}
\usepackage{booktabs}
\usepackage{amsmath}
\usepackage{amssymb}
\usepackage{xcolor}
\usepackage{tikz}
\usepackage{pgfplots}
\usepackage{multirow}
\usepackage{pgfplots}
\usepackage{array}
\pgfplotsset{compat=1.18}
\usetikzlibrary{arrows.meta,positioning,calc}

\definecolor{bsPanelBg}{HTML}{0D0F14}
\definecolor{bsInk}{HTML}{AEB8C8}
\definecolor{bsCmd}{HTML}{E8EEF7}
\definecolor{bsWoodPan}{HTML}{B5763F}
\definecolor{bsWoodPanR}{HTML}{9C6231}
\definecolor{bsWallpaper}{HTML}{F06292}
\definecolor{bsPlaster}{HTML}{CF9226}
\definecolor{bsDoor}{HTML}{F7C59F}
\definecolor{bsWindow}{HTML}{53F4FF}
\definecolor{bsTile}{HTML}{43BD43}
\definecolor{bsWood}{HTML}{C48545}

\newcommand{\bsswatch}[1]{%
  {\setlength{\fboxsep}{0pt}%
   \setlength{\fboxrule}{0.3pt}%
   \fcolorbox{black!45}{#1}{%
     \phantom{\rule{1.6ex}{1.6ex}}%
   }}%
}

\newcommand{\gtpanel}[2]{%
  \begin{tikzpicture}[inner sep=0.1pt, outer sep=0.1pt]
    \node (img) {\includegraphics[width=0.41\linewidth]{#1}};
    \node[anchor=north west, font=\scriptsize\bfseries, text=black!80,
          fill=white, fill opacity=0.75, text opacity=1, inner sep=2pt,
          rounded corners=1pt]
      at ([xshift=1.5mm, yshift=-1.5mm]img.north west) {#2};
  \end{tikzpicture}}
  
\usepackage[accsupp]{axessibility}  % Improves PDF readability for those with disabilities.

\usepackage[pagebackref,breaklinks,colorlinks,citecolor=eccvblue]{hyperref}
\usepackage{orcidlink}
\usepackage{placeins}
\usepackage{capt-of}

\newcommand{\code}[1]{{\small\texttt{#1}}}

\begin{document}

\title{BIMScript: Material-Aware Structured Scene Programs for BIM Ingestion}
\titlerunning{BIMScript}

% \title{\smash{\raisebox{-1.55em}{\includegraphics[height=2.5em {figures/bimscript_icon.pdf}}} \hspace{0.1em}
% BIMScript: Material-Aware Structured Scene Programs for BIM Ingestion}

\author{Prakash Naikade\inst{1,2,3}\orcidlink{0009-0000-5901-8975} \and
Thomas Moeslund\inst{1,2,3}\orcidlink{0000-0001-7584-5209} \and
Andreas M\o gelmose \inst{1,2,3}\orcidlink{0000-0003-0328-382X}}

\authorrunning{P.~Naikade et al.}

\institute{AI:Xpertise Lab, Aalborg University, Denmark \and
Visual Analysis and Perception Laboratory, Aalborg University, Denmark \and
Pioneer Centre for Artificial Intelligence, Denmark\\[2pt]
\centerline{\small\url{https://BIMScriptWorld.github.io/BIMScript/}}}

\maketitle
\vspace{-4mm}

% ============================================================
% Teaser: placed between the title block and the abstract.
% Non-floating so it stays exactly here; \captionof supplies the number.
% ============================================================
\begin{center}
\resizebox{\textwidth}{!}{%
\begin{tikzpicture}[
  card/.style={rounded corners=4pt, draw=black!25, line width=0.4pt},
  arr/.style={-{Stealth[length=2.4mm]}, line width=1.0pt, black!45},
  lab/.style={font=\bfseries\scriptsize, align=center, text=black!85, anchor=south},
]
% ---------------------------------------------------------------
% One content band, y in [0, 2.90], for every column, and one gap
% (0.70) between all four columns. Each image is sized to exactly fill
% its card and clipped with the card's own rounded corners, so no
% content is cropped and no square corner escapes the border.
% ---------------------------------------------------------------
\coordinate (origin) at (0,0);

% ---- Column 1: point cloud over an egocentric keyframe strip ----
\begin{scope}
  \clip[rounded corners=4pt] (0.06,0.85) rectangle (2.24,2.90);
  \node[anchor=south west, inner sep=0pt] (pc) at (0.06,0.85)
    {\includegraphics[height=2.06cm]{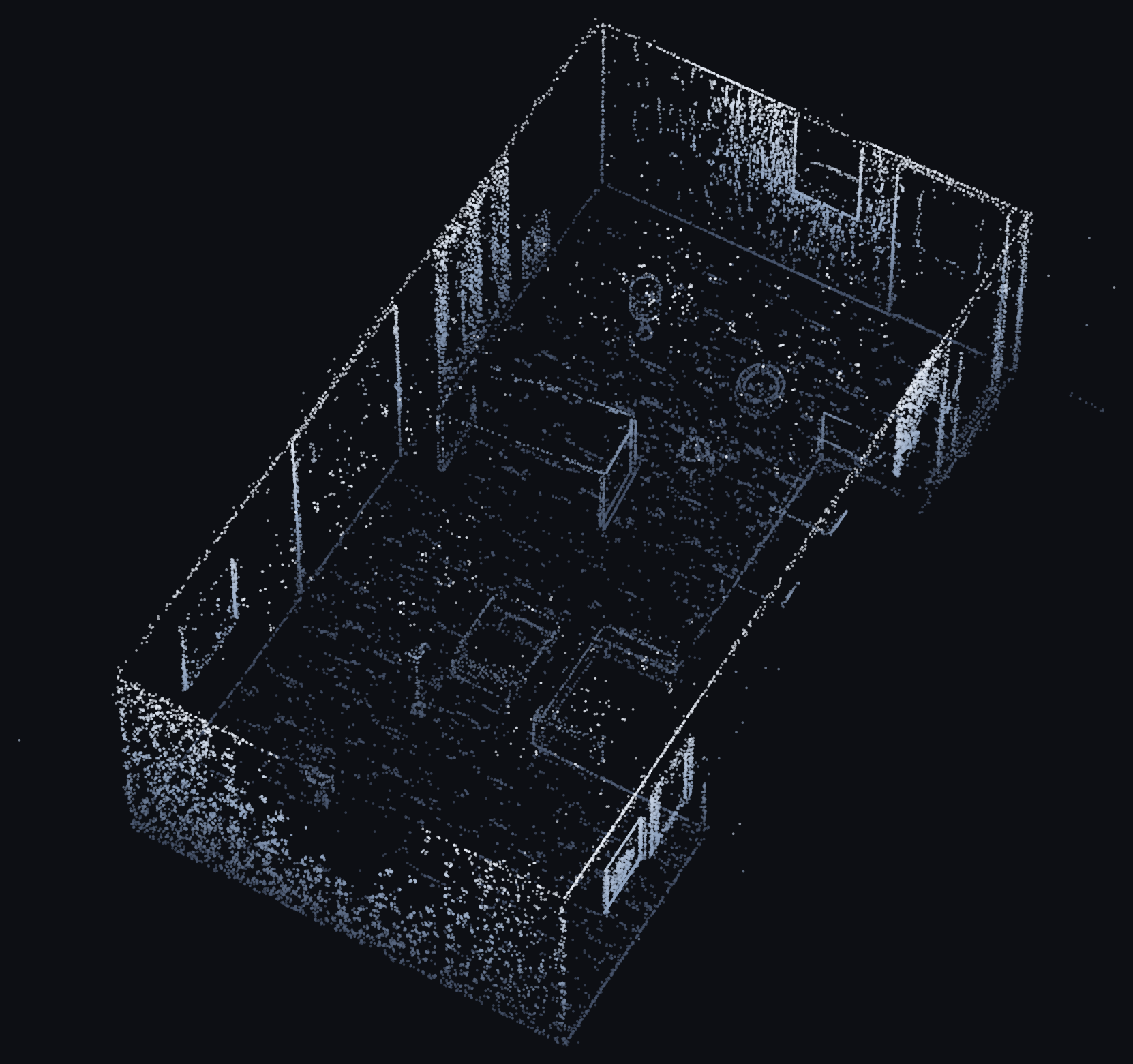}};
\end{scope}
\draw[card] (0.06,0.85) rectangle (2.24,2.90);

\foreach \i/\x in {frame_0/0.05, frame_1/0.80, frame_2/1.55} {
  \begin{scope}
    \clip[rounded corners=2pt] (\x,0) rectangle (\x+0.70,0.70);
    \node[anchor=south west, inner sep=0pt] at (\x,0)
      {\includegraphics[width=0.70cm]{figures/teaser/\i.jpg}};
  \end{scope}
  \draw[card, rounded corners=2pt] (\x,0) rectangle (\x+0.70,0.70);
}

% ---- Column 2: encoder-decoder model (gap 0.70 after column 1) ----
\node[card, fill=bsPanelBg, minimum width=1.8cm, minimum height=2.0cm]
  (model) at (3.90,1.80) {};
\foreach \i/\y in {1/-0.40, 2/0, 3/0.40} {\coordinate (l\i) at (3.45,1.80+\y);}
\foreach \i/\y in {1/-0.60, 2/-0.20, 3/0.20, 4/0.60} {\coordinate (m\i) at (3.90,1.80+\y);}
\foreach \i/\y in {1/-0.40, 2/0, 3/0.40} {\coordinate (r\i) at (4.35,1.80+\y);}
\foreach \a in {1,2,3} \foreach \b in {1,2,3,4}
  \draw[bsInk!55, line width=0.4pt] (l\a) -- (m\b);
\foreach \a in {1,2,3,4} \foreach \b in {1,2,3}
  \draw[bsInk!55, line width=0.4pt] (m\a) -- (r\b);
\foreach \p in {l1,l2,l3,m1,m2,m3,m4,r1,r2,r3} \fill[bsCmd] (\p) circle (0.068);

% ---- Column 3: decoded program (gap 0.70 after column 2) ----
\node[card, fill=bsPanelBg, anchor=west, align=left, inner sep=6pt,
      font=\ttfamily\scriptsize] (progn) at (5.50,1.80) {%
  {\color{bsCmd}make\_wall}~{\color{bsInk}id=0, ...,}\\
  {\color{bsInk}~~}{\color{bsWoodPan}material=wood\_paneling}{\color{bsInk}, condition=good}\\
  {\color{bsCmd}make\_wall}~{\color{bsInk}id=1, ..., }{\color{bsWallpaper}material=wallpaper}\\
  {\color{bsCmd}make\_wall}~{\color{bsInk}..., }{\color{bsPlaster}material=painted\_plaster}\\
  {\color{bsDoor}make\_door}~{\color{bsInk}wall=0, ..., material=composite}\\
  {\color{bsWindow}make\_window}~{\color{bsInk}wall=2, ..., material=wood}%
};

% ---- Legend: beneath columns 2 and 3, inside the same band ----
\coordinate (legc) at ($(model.west)!0.5!(progn.east)$);
\coordinate (legy) at (0,0.28);
\node[font=\scriptsize, text=black!90, anchor=center] at (legc |- legy) {%
  \bsswatch{bsWoodPanR}~wood paneling \,
  \bsswatch{bsWallpaper}~wallpaper \,
  \bsswatch{bsPlaster}~painted plaster \,
  \bsswatch{bsDoor}~door \,
  \bsswatch{bsWindow}~window};

% ---- Column 4: 3D BIM, full band height (gap 0.70 after column 3) ----
\coordinate (c4) at ([xshift=0.70cm]progn.east |- origin);
\begin{scope}
  \clip[rounded corners=4pt] (c4) rectangle ($(c4)+(2.94,2.90)$);
  \node[anchor=south west, inner sep=0pt] at (c4)
    {\includegraphics[height=2.90cm]{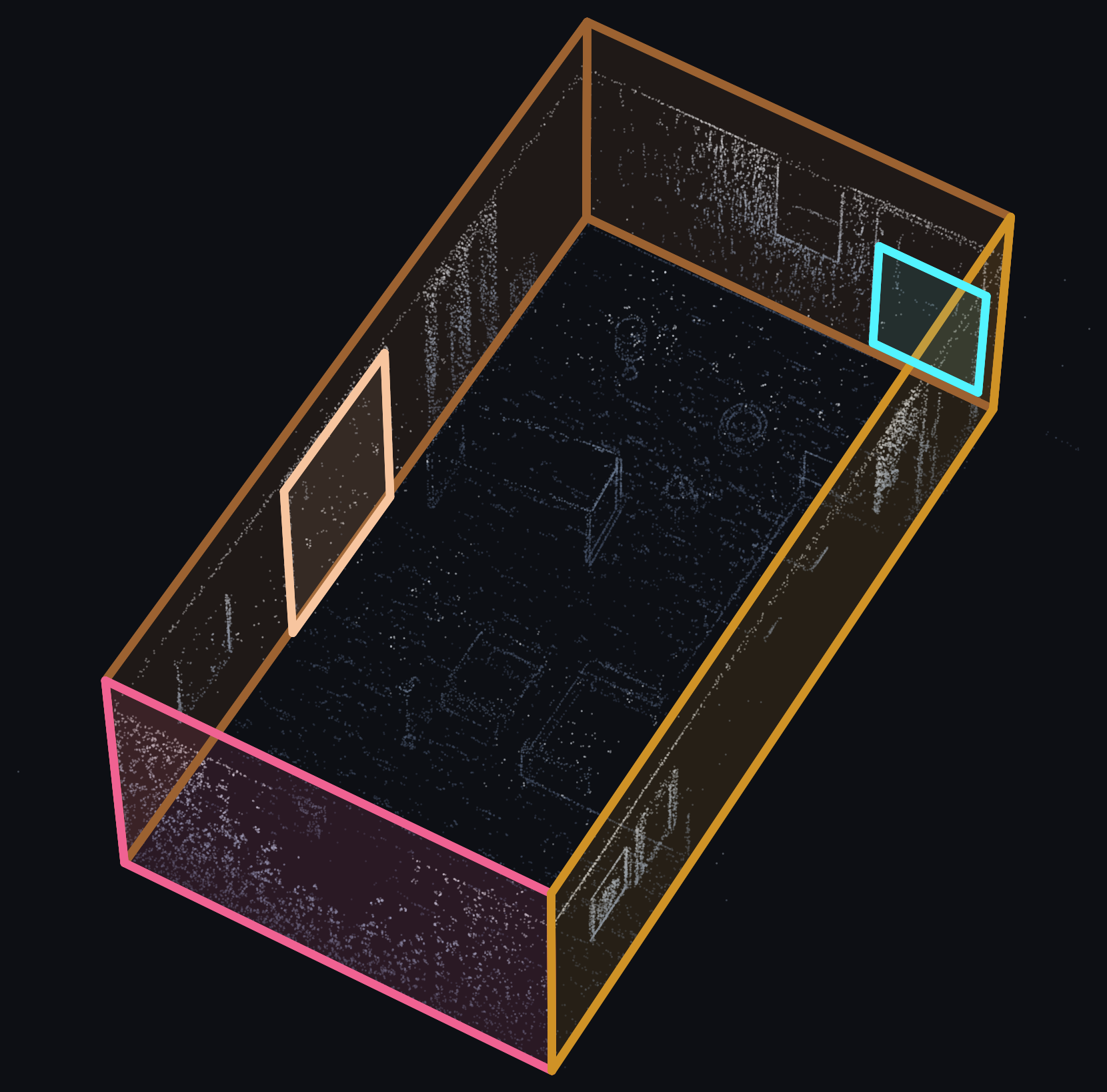}};
\end{scope}
\draw[card] (c4) rectangle ($(c4)+(2.94,2.90)$);

% ---- Flow ----
\draw[arr] (2.25,1.8) -- (model.west);
\draw[arr] (2.18,0.31) to[out=0,in=205] ([yshift=-3mm]model.west);
\draw[arr] (model.east) -- (progn.west);
\draw[arr] (progn.east) -- ($(c4)+(0,1.8)$);

% ---- Titles: one line each, on one row ----
\node[lab] at (1.15,3.02) {Scan + Images};
\node[lab] at (3.90,3.02) {BIMScript};
\coordinate (top) at (0,3.02);
\node[lab] at (progn |- top) {Material- and Condition-aware Program};
\node[lab] at ($(c4)+(1.47,3.02)$) {Predicted 3D BIM};
\end{tikzpicture}}
{\captionsetup{hypcap=false}
\captionof{figure}{\textbf{BIMScript turns a scan into a material- and condition-aware BIM program.} A point cloud and its egocentric keyframes are
encoded jointly; the decoder emits a structured program in which every element
carries geometry, material, and condition, and each command maps one-to-one
onto a native Revit/IFC object.}
\label{fig:bimscript_teaser}}
\end{center}
\vspace{-6mm}

\begin{abstract}
Structured-language models such as SceneScript reconstruct a scene as a short program of parametric commands, an inherently editable and semantically explicit representation. We ask three questions that stand between such models and their most compelling application, automated ingestion of existing buildings into BIM tools, studied here on synthetic scans: \emph{what} is the scene made of, \emph{how fast} can it be produced, and \emph{exactly where} is each element. BIMScript answers all three within one grammar. First, we extend the layout language with per-element \emph{material} and \emph{condition} attributes, supervised by a vision-language-model material-passport corpus we build over 100k synthetic scenes (1.9M pseudo-labeled elements), and route image appearance to the material tokens through a lifted-feature point encoder. Second, we show that autoregressive decoding of these programs is dominated not by compute but by kernel-launch and host-synchronization overhead, and remove it with an output-exact CUDA-graph decoder (1.9 vs 6.4\,ms/step, $3.4\times$) plus a grammar-parallel, tolerance-verified draft-and-verify scheme that exploits the deterministic entity schema. Third, we address the model's 5cm token-grid granularity with training-free geometric snapping and a hybrid discrete--continuous decoder head that regresses a sub-bin offset, and measure how much of the residual error each recovers. Because each command maps one-to-one onto a native Revit object, we validate direct ingestion into a BIM authoring tool end to end with a working add-in and its IFC4 export, and the same program's language form is designed to support LLM-driven, sustainability-aware reasoning over the built asset.

\keywords{3D scene understanding \and Structured language models \and \\
Efficient decoding \and Material recognition \and Building Information Modeling }
\end{abstract}

% =====================================================================
\section{Introduction}
\label{sec:intro}

The built environment is responsible for a large share of global material consumption and carbon emissions, and the overwhelming majority of the buildings that will exist in 2050 already stand today \cite{unep_2023, unhabitat_2023}. Decarbonizing, renovating, re-valuing, or eventually disassembling these assets increasingly relies on a digital model of \emph{what is already there} \cite{BYERS_review_sustrch_2024, DeWolf_D5_circular_workflow_2024}. In practice such models are Building Information Models (BIM): parametric, object-level descriptions of walls, openings, slabs, floors and their properties, consumed by tools such as Autodesk Revit and exchanged through open standards like IFC~\cite{iso16739}. Producing a BIM of an existing building, however, remains a slow, largely manual ``scan-to-BIM'' process~\cite{volk2014building,tang2010automatic,macher2017point}: a surveyor captures a point cloud, and a modeler hand-traces geometry and hand-labels properties element by element.

Structured-language models offer an appealing shortcut. SceneScript~\cite{avetisyan2024scenescript} reconstructs a scene not as a mesh or an occupancy grid but as a short \emph{program}, a sequence of parametric commands such as \code{make\_wall} and \code{make\_door}, emitted autoregressively by a transformer decoder conditioned on an encoded point cloud. The representation is compact, human-readable, directly editable, and, crucially, already shaped like the object model a BIM authoring tool expects. Yet three gaps separate this elegant formulation from practical as-built BIM ingestion, and they map onto \emph{three questions}: (1) \emph{What is in the scene}? SceneScript predicts geometry and coarse class, but a BIM element is defined as much by its \emph{material} and \emph{condition} as by its shape. A wall is not merely a wall: it is a \emph{brick} wall in \emph{good} condition, and that attribute drives every downstream sustainability workflow. The layout grammar carries no such information. (2) \emph{Fast}? Autoregressive decoding of one scene program takes on the order of seconds on a modern GPU. For interactive modeling and for ingesting building stock at portfolio scale, this is the binding constraint. (3) \emph{Exactly where}? SceneScript itself notes that high-level commands ``can be challenging to capture fine-grained geometric details with extremely high precision (i.e. mm)''. In the reference configuration this is not a vague limitation but an exact one: coordinates are discretized onto a 5cm grid, so nearest-bin quantization alone has a worst-case error of $\pm2.5$\,cm before any model error, below the tolerance BIM level-of-accuracy (LOA) and level-of-development (LOD) specifications demand~\cite{usibd2019loa, EleannaReliabilityInBIM, Abualdenien_LOD_validation_2020}.
% , level-of-geometry (LOG)

BIMScript answers all three questions within a single, backward-compatible grammar, and is designed around the payoff: each command corresponds one-to-one with a native Revit/IFC object (\cref{sec:application}), so the emitted program is positioned to ingest directly into BIM authoring tools; and because the scene is expressed in language, it becomes a first-class input to large language models (LLMs) for downstream, sustainability-aware Architecture, Engineering, and Construction (AEC) reasoning: real-estate valuation, life-cycle assessment (LCA), adaptive-reuse and disassembly planning, and circular-economy waste reduction.

Our key contributions are:
\begin{itemize}
\item \emph{What}: We propose BIMScript, a material- and condition-aware scene language, and build a material-passport corpus of 1{,}900{,}908 pseudo-labeled
elements over ${\approx}100$k scenes to supervise it; a lifted-feature point encoder routes image appearance to the material tokens. %, its contribution isolated by a matched-population ablation.
\item \emph{Fast}: We profile decoding, show that ${\sim}65\%$ of it is kernel-launch and host-synchronization overhead rather than compute, and remove that overhead with an output-exact CUDA-graph decoder ($3.4\times$), with a grammar-parallel, tolerance-verified draft-and-verify scheme that exploits the deterministic entity schema.
\item \emph{Exactly where}: We add training-free geometric snapping and a hybrid head that regresses a bounded within-bin offset, with the 5cm grid's contribution to coordinate error measured directly.
\item Lastly, we publish a public-artifact reproduction audit of SceneScript, and a working BIMScript-to-Revit adapter that creates native elements carrying material and condition parameters.
\end{itemize}

% removed following para 
% All extensions sit behind configuration flags that default to legacy behavior, so existing layout checkpoints and metrics are unchanged unless a feature is enabled; \cref{fig:arch} summarizes the system.

% =====================================================================
\section{Related Work}
\label{sec:related}

\subsection{Floorplan reconstruction and scan-to-BIM.}
A long line of work reconstructs floorplans and layouts from scans, e.g.\
{FloorNet}~\cite{liu2018floornetunifiedframeworkfloorplan},
{MonteFloor}~\cite{stekovic2021montefloorextendingmctsreconstructing},
{RoomFormer}~\cite{yue2023connectingdotsfloorplanreconstruction},
{PolyRoom}~\cite{liu2024polyroomroomawaretransformerfloorplan},
{FRI-Net}~\cite{xu2024frinetfloorplanreconstructionroomwise} and
{CAGE}~\cite{liu2025cagecontinuityawareedgenetwork}, or aligns CAD models to
RGB-D scans~\cite{avetisyan2020scenecad}; large scan and layout datasets
underpin them \cite{avetisyan2024scenescript,procthor,dai2017scannet,armeni2017joint2d3dsemanticdataindoor,zheng2020structured3dlargephotorealisticdataset,bieri2025houselayout3d,abouagour2025resplanlargescalevectorgraphdataset,ganon2024wafflemultimodalfloorplanunderstanding}.
In the AEC community, scan-to-BIM is typically a multi-stage geometric-fitting pipeline \cite{tang2010automatic,macher2017point,volk2014building}. These methods target geometry and topology; material and condition are added manually afterwards. BIMScript predicts geometry, material, and condition jointly in one autoregressive pass and emits a program that is already an object model.

\subsection{Structured-language scene models.}
SceneScript~\cite{avetisyan2024scenescript} recasts layout estimation as
next-token prediction over a domain-specific command language, following a broader line of work that treats geometry as language: PolyGen for meshes~\cite{nash2020polygen}, DeepCAD~\cite{wu2021deepcad} and CAD-as-Language~\cite{ganin2021computer} for parametric CAD. These inherit the transformer decoder~\cite{vaswani2017attention} and its autoregressive cost. Fast SceneScript (FSS)~\cite{yin2025fastscenescript} accelerates SceneScript with multi-token prediction (MTP) and self-speculative decoding, reducing iteration count at a small quality cost. BIMScript differs on all three axes: we add \emph{semantics} (materials) rather than only geometry, we attack decode \emph{overhead} before iteration count (an orthogonal and multiplicative gain), and we replace generic MTP with a \emph{grammar-parallel} scheme that is exact by construction because the entity schema makes future token \emph{types} deterministic. Xie \etal~\cite{xie2025localcorrections} also bring NLP-style \emph{infilling} to SceneScript, but for a different problem and by a different mechanism: their human-in-the-loop local correction rearranges the sequence FIM-style and regenerates the masked entities \emph{token by token}, trained from scratch with pose anchoring. Our \code{[INFILL]} embedding instead masks slots \emph{in place} and uses the schema-determined slot types to draft a whole entity in a \emph{single} verified forward pass. For them infilling is the task; for us it is an acceleration primitive added to an existing checkpoint by a short fine-tune.

\subsection{Efficient autoregressive decoding.}
Speculative decoding~\cite{leviathan2023fast,chen2023accelerating}, multi-token prediction~\cite{gloeckle2024better,cai2024medusa}, blockwise parallel decoding \cite{stern2018blockwise}, KV-caching and multi-query attention \cite{shazeer2019fast}, and CUDA graphs~\cite{gray2019cudagraphs} all reduce transformer inference cost. Most target large language models where each step is compute-heavy. SceneScript's decoder is tiny (a step through 4 layers of width
$d\,{=}\,512$ is $\sim$2\,MFLOPs), so the dominant cost is overhead, which
reframes which of these tools matter, as we show in \cref{sec:fast}.

\subsection{Material passports, BIM, digital twins, and circular economy.}
Material recognition from images is well studied
\cite{bell2015material,bell2013opensurfaces,upchurch2022dense,schwartz2020recognizing}, but almost always as 2D image labeling. Connecting per-pixel or per-region material evidence to \emph{3D building elements} in a reconstruction, at scale and without per-scene manual annotation, is the gap BIMScript's material-passport pipeline fills, using a vision-language model~\cite{gemmateam2025gemma3} to distill appearance into element-level labels.

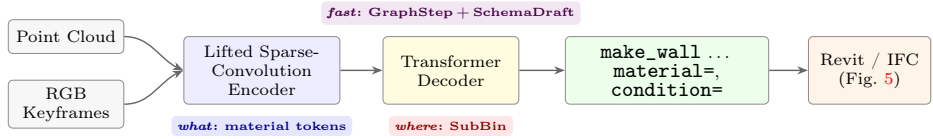
\begin{figure}[tb]
\centering
\resizebox{\linewidth}{!}{%
\begin{tikzpicture}[
  font=\scriptsize,
  box_small/.style={draw=black!45, rounded corners=2pt, align=center, inner sep=3pt,
              minimum height=5mm},
  box/.style={draw=black!45, rounded corners=2pt, align=center, inner sep=3pt,
              minimum height=10mm},
  tag/.style={draw=none, fill=#1!10, rounded corners=2pt, align=center,
              inner sep=2.5pt, font=\tiny\bfseries, text=#1!55!black},
  arr/.style={-{Stealth[length=1.6mm]}, semithick, black!60},
]
\node[box_small, fill=black!3, text width=15mm] (pc)   at (0,0.5)  {Point Cloud};
\node[box_small, fill=black!3, text width=15mm] (rgb)  at (0,-0.5) {RGB Keyframes};
\node[box, fill=blue!7,  text width=21mm] (enc)  at (2.9,0)  {Lifted Sparse-Convolution Encoder};
\node[box, fill=yellow!15, text width=18mm] (dec)  at (5.7,0)  {Transformer\\Decoder};
\node[box, fill=green!8, text width=28mm] (prog) at (8.9,0)
  {\code{make\_wall} \dots\\\code{material=}, \code{condition=}};
\node[box, fill=orange!9, text width=16mm] (bim) at (11.9,0) {Revit / IFC\\(\cref{fig:revit})};

\draw[arr] (pc.east)  to[out=0,in=155] (enc.west);
\draw[arr] (rgb.east) to[out=0,in=205] (enc.west);
\draw[arr] (enc) -- (dec);
\draw[arr] (dec) -- (prog);
\draw[arr] (prog) -- (bim);

\node[tag=blue,   below=1.6mm of enc] {\emph{what}: material tokens};
\node[tag=violet, above=1.6mm of dec] {\emph{fast}: GraphStep\,+\,SchemaDraft};
\node[tag=red,    below=1.6mm of dec] {\emph{where}: SubBin};
\end{tikzpicture}}
\caption{BIMScript overview. A lifted-feature encoder routes image appearance
into the point encoder, so the decoder can condition each material and
condition token on the element's own geometry tokens. Decoding is accelerated
by \textit{GraphStep} and \textit{SchemaDraft} and refined below the token grid by \textit{SubBin}
(\cref{fig:fastbim}), and every emitted command maps one-to-one onto a native
Revit/IFC object.}
\label{fig:arch}
\end{figure}

Material passports~\cite{luscuere2017materials,honic2019improving} and
circularity indicators~\cite{heisel2020calculation,pomponi2017circular} require element-level material inventories; BIM-based LCA~\cite{soust2017critical} requires geometry and materials in a machine-readable model. BIMScript (\cref{fig:arch}) produces exactly this inventory automatically. A broader survey of XR/AI for AEC situates this direction~\cite{naikade2026xrai}.

% =====================================================================
\section{Background and a Reproducible Baseline}
\label{sec:background}

\subsection{SceneScript in brief}
SceneScript encodes a point cloud with a sparse-convolutional
encoder~\cite{tang2022torchsparse,choy20194d} and decodes a command program with a transformer decoder attending to the encoded context. Continuous parameters (coordinates, sizes) are discretized into $B$ bins over a fixed world range and predicted as categorical tokens; a small type finite-state machine (FSM) constrains which token kind is legal at each step. In the configuration matching the public checkpoint the world range is $[0,32]$\,m and $B{=}640$, giving a $32/640 = 5$cm grid, the paper's stated point cloud resolution.

\subsection{Compatibility and target-coverage issues} % exposed by public artifacts
\label{sec:compat}
Naive training does not reproduce the public checkpoint's behavior, for two reasons worth recording for any follow-up work on this family. First, a \emph{discretization/reconstruction mismatch}: the tokenizer pairs a rounding rule at discretization with a reconstruction rule at decode, and pairing floor-rounding with bin-edge reconstruction biases every coordinate by up to a full bin, which alone pushes wall-corner error past the F1@5cm threshold irrespective of model quality. Second, a \emph{translation-frame} issue: anchoring the coordinate frame at the observed point-cloud minimum places ground-truth geometry of \emph{unobserved} floor area at negative coordinates, which the tokenizer then discards, a silent recall loss of 46{,}218 removal events in one training run (\cref{app:repro}). We adopt the unbiased pairing (round at discretize, edge at reconstruct) and a margin-anchored frame with a bounded clamp, applied identically at train and inference.

\Cref{app:repro} gives both derivations, the controlled A/B that identifies
which convention is consistent with the released weights, the measured
round-trip oracle, and the mitigation's residual cost ($0.6\%$ of entities at
evaluation time), rather than asserting either issue is fully resolved.

%%------table 1 : proxy of scenescript paper table 2 -----%%
\begin{table}[tb]
\caption{Reproducing the SceneScript point-only encoder--decoder from public artifacts for layout estimation. \textit{Top}: 1{,}000 held-out test scenes, same evaluator, one greedy decode per model, so F1@5cm and both AvgF1 columns come from a single set of predictions. ``Local-FSS'' is our reimplementation of Fast SceneScript's 20-threshold average (\cref{sec:parity}); parameters are measured from the instantiated models. \textit{Bottom}: numbers as published, on each paper's own split and evaluator, neither of them public.} 
% FSS reports decoder-only parameter counts; we report their SSD $n{=}10$ variant.
% and not row-comparable to the top block, and we cannot verify their parameter counts.}
\label{tab:repro}
\centering
\small
\setlength{\tabcolsep}{3pt} 
\setlength{\aboverulesep}{0pt}     
\setlength{\belowrulesep}{0pt} 
\begin{tabular}{@{}l|ccc|c|c|c@{}}
\multirow{2}{*}{Model} & \multicolumn{3}{c|}{Params (M) $\downarrow$}
  & \multirow{2}{*}{F1@5cm $\uparrow$} & \multirow{2}{*}{AvgF1 $\uparrow$}
  & \multirow{2}{*}{\shortstack{AvgF1\\(local-FSS) $\uparrow$}} \\
 & Enc. & Dec. & Total & & & \\
\midrule
Public ckpt          & 10.47 & 15.38 & 25.85 & 0.581 & 0.667 & 0.931 \\
Ours (640-bin grid)        & 20.15 & 36.42 & 56.57 & 0.554 & 0.658 & 0.930 \\
Ours (2048-bin grid)       & 20.15 & 37.86 & 58.01 & 0.570 & 0.672 & 0.896 \\
\midrule
SceneScript Table\,2       & $\approx20$    & $\approx35$ & $\approx55$ & 0.848 & 0.784 & -- \\
FSS Table\,1 (SSD,  $n{=}10$) & --    & 15.05     & -- & --    & --    & 0.912 \\
\end{tabular}
\end{table}

\subsection{Reproduction parity, and what the token grid buys}
\label{sec:parity}

\noindent\textbf{What public artifacts can support.} SceneScript's published Table~2 (mean F1@5cm $0.848$) cannot be audited from public artifacts. The released checkpoint's value-embedding matrix has $646{=}640{+}6$ rows and 2{,}000 learned positions, whereas the paper states a vocabulary of 2{,}048, so the released artifact does not match the paper-described architecture; the dataset splits are not public; and Fast SceneScript, which retrains the same architecture, reports no F1@5cm at all. Concretely, the released weights score $0.581$ mean F1@5cm on our test split against the paper's $0.848$ on its own (non-public) split; we cannot attribute that difference, so we treat the artifact, not the published table, as the
reproduction target. The released checkpoint is therefore the only verifiable anchor, and we anchor every claim to it on fixed splits of our own. Two controls then separate evaluator validity from training validity. For the \emph{evaluator}, our reimplementation of an FSS-style threshold protocol scores the released checkpoint $0.931$ AvgF1, against the $0.896$ Fast SceneScript reports for the same weights on their different split. For \emph{training}, on identical scenes with a single decode per scene, our retrained model reaches $0.930$ local-FSS AvgF1 against the released checkpoint's $0.931$ (\cref{tab:repro}). We read this as evidence that our training path reaches the released artifact's behavior, not as a general equivalence claim. Our runs carry $2.19\times$ (640-bin) and $2.24\times$ (2{,}048-bin) the released checkpoint's parameters, but both sit within $3\%$ of the ${\approx}55$M SceneScript specifies, so the gap is the released artifact being roughly half the size of its own paper's architecture rather than our models being oversized. The same predictions yield $0.667$ under our dense-threshold protocol and $0.931$ under the local FSS-style one, so ``mean F1'' is protocol-dependent: we report both and never compare across them.

\noindent\textbf{The token grid is a modest, two-sided knob.} A 300-scene round-trip through our tokenizer (\cref{app:repro}) shows the representation is not the binding cap: ground truth reconstructed through the tokenizer scores $0.994$ F1@5cm at 640 bins, and frame-boundary censoring removes only $0.6\%$ of entities. The residual error is instead \emph{amplified} by grid pitch, since on a 640-bin (5cm) grid a $\pm1$-bin error is already a 5--10\,cm miss while on a 2{,}048-bin (1.56\,cm) grid the same error passes. We therefore decouple the \emph{language} grid from the point-cloud voxel grid, leaving the input at 640 bins while the value vocabulary becomes an independent knob, and train a matched-budget 2{,}048-bin run. The effect is real but small and cuts both ways: strict F1@5cm rises $0.554\!\to\!0.570$ while the coarse local-FSS average falls $0.930\!\to\!0.896$ (\cref{tab:repro}). A finer grid therefore buys strict-threshold precision and gives back some coarse-threshold
robustness. We report it as a design trade-off rather than as the explanation
for the distance to the published number, and keep both grids in play in what
follows. Training curves and a recipe negative result are in
\cref{app:repro}.

% =====================================================================
\section{BIMScript: What Is in the Scene}
\label{sec:what}

\subsection{Material-aware scene language}
We keep SceneScript's flat, definition-driven grammar and extend the parameter
list of each building element with two inline categorical attributes:
{\small \centering
\begin{verbatim}
make_wall,  id=0, a_x=.., .., b_z=.., height=..,
            thickness=0.0, material=painted_plaster, condition=good
make_door,  id=1000, wall0_id=0, position_x=.., .., width=..,
            height=.., material=composite, condition=new
make_window,id=2000, wall0_id=0, position_x=.., .., width=..,
            height=.., material=aluminum, condition=good
\end{verbatim}}

\noindent Inline attributes ride the existing categorical value-token space (the
class index lives below the numeric-bin range, so the vocabulary does not grow),
require no decoder change, and, decisively, condition each material token on
the element's \emph{own} geometry tokens through self-attention. We deliberately
avoid a separate \code{set\_material,\,target\_id=..} command block: decoded
element ids are positional and regenerated at decode time, so cross-referencing a
separately predicted attribute to its element is brittle, and it roughly doubles
sequence length. Materials use atomic \code{snake\_case} labels from a closed
taxonomy; the two attributes add $\sim$2 tokens per element ($\sim$40 per scene),
comfortably within the 2048-token budget. The taxonomies are frozen and
append-only, so the type-token enumeration extends the layout enumeration without
renumbering: every layout checkpoint stays valid and material models warm-start
from one.

% better abbrivation for improvement steps to use subsequently in paper figure

\subsection{A material-passport corpus at scale}
\label{sec:passport}
Aria Synthetic Environment (ASE)~\cite{avetisyan2024scenescript, engel2023aria} dataset's synthetic indoor scenes give exact ground-truth \emph{geometry} but no per-element material labels. We generate them with a per-scene \emph{material passport}: for every element we gather the RGB observations that project onto it and prompt a vision-language model~\cite{gemmateam2025gemma3} on the resulting crops for a material and condition label. Two further cues, a dense material-segmentation predictor and classical color/texture statistics, are computed independently and stored beside it. They do not supply the label; they measure how far it can be trusted (\cref{app:material}). Geometry lines are copied verbatim from ground truth, so layout supervision is byte-identical to the layout task.

The corpus (\cref{tab:corpus}) has 99{,}990 scenes and 1{,}900{,}908 elements over a 14-material, 5-condition taxonomy, with zero join failures.
The full material and condition taxonomies are
$[$\,\code{aluminum},\allowbreak{}
\code{brick},\allowbreak{}
\code{composite},\allowbreak{}
\code{concrete},\allowbreak{}
\code{glass},\allowbreak{}
\code{metal},\allowbreak{}
\code{painted\_plaster},\allowbreak{}
\code{pvc},\allowbreak{}
\code{stone},\allowbreak{}
\code{tile},\allowbreak{}
\code{wallpaper},\allowbreak{}
\code{wood},\allowbreak{}
\code{wood\_paneling},\allowbreak{}
\code{unknown}$]$ and
$[$\code{new},\allowbreak{}
\code{good},\allowbreak{}
\code{worn},\allowbreak{}
\code{damaged},\allowbreak{}
\code{unknown}\,$]$, respectively.
Two properties matter for honest evaluation. The distribution is heavily imbalanced (\code{good} covers 83\% of conditions, \code{damaged} just 0.05\%), so we report macro-F1 alongside accuracy. And the pseudo-labels are noisy: the VLM agrees with the classical-CV cue on only ${\sim}25\%$ of elements and with the dense-segmentation cue on ${\sim}34\%$. That low agreement is evidence that materials are genuinely hard to pin down from any single cue, so some of the residual error in \cref{tab:material} belongs to the labels rather than to BIMScript. It is not a formal bound on achievable accuracy.

\begin{figure}[tb]
\centering
\setlength{\tabcolsep}{0.5pt}
\begin{tabular}{cc}
\footnotesize Ground Truth & \footnotesize BIMScript Prediction \\
\includegraphics[width=0.48\linewidth]{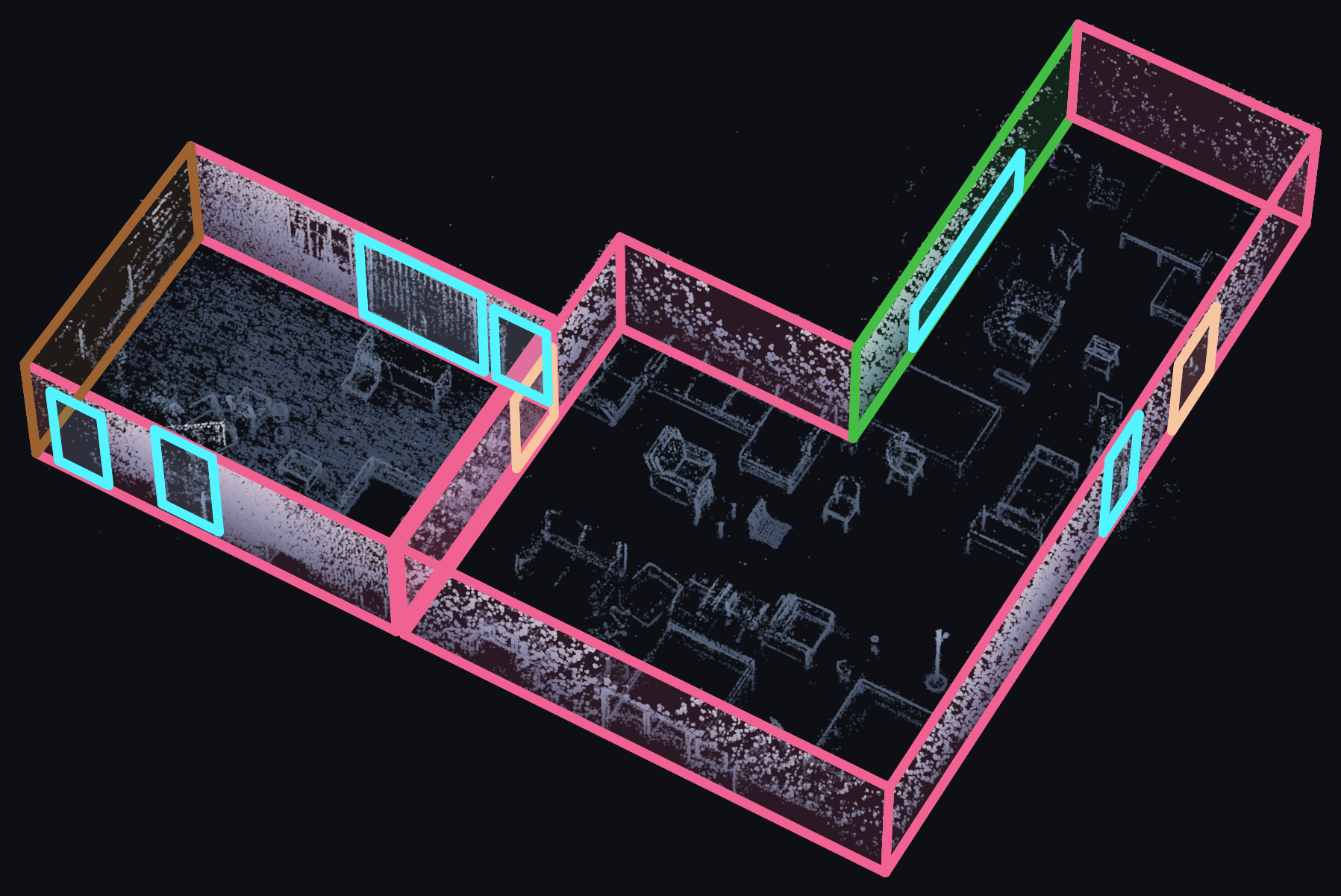} &
\includegraphics[width=0.48\linewidth]{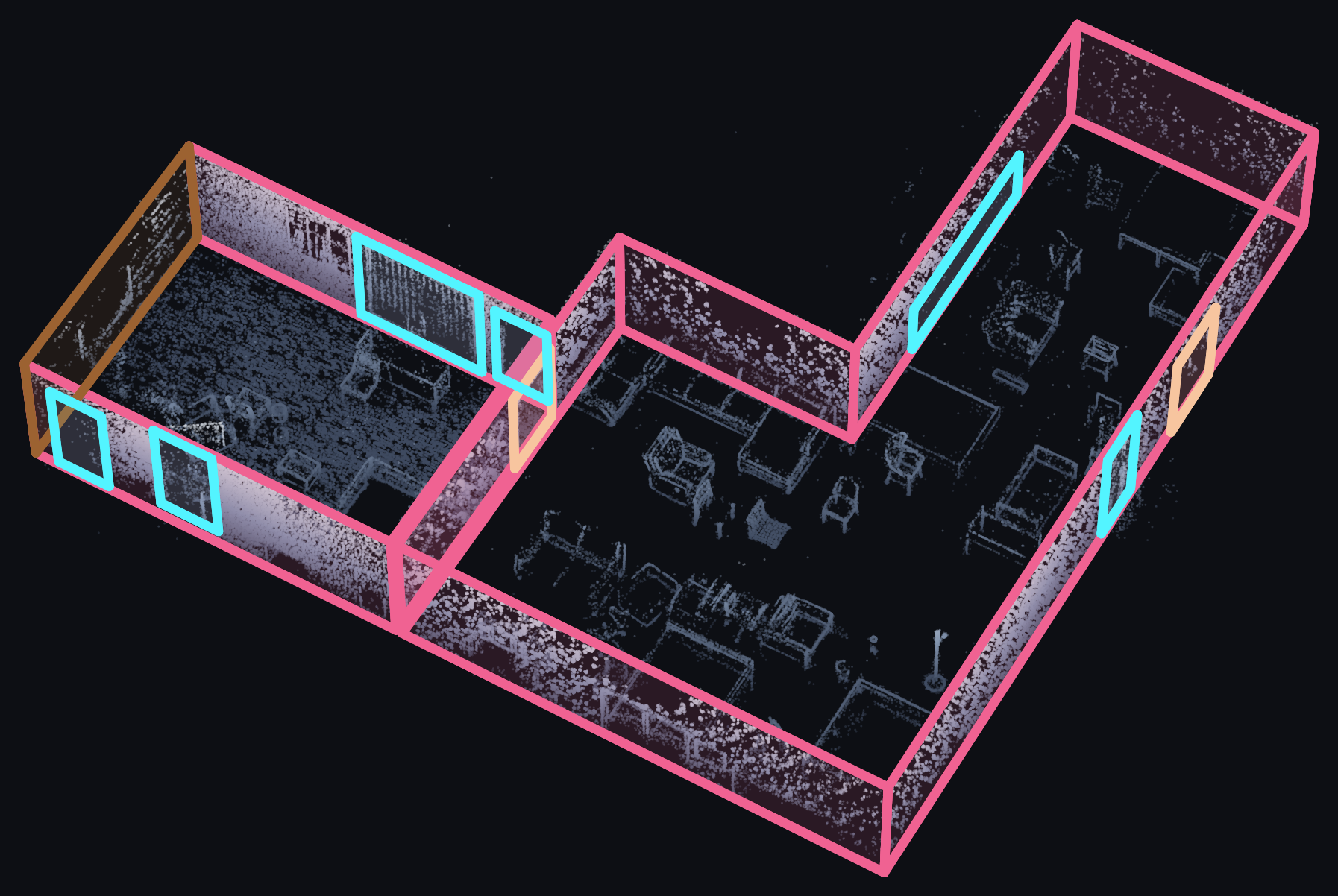} \\[2pt]
\multicolumn{2}{c}{\scriptsize
  \bsswatch{bsWallpaper}~wallpaper\,
  \bsswatch{bsTile}~tile\,
  \bsswatch{bsWoodPanR}~wood paneling\,
  \bsswatch{bsDoor}~door\,
  \bsswatch{bsWindow}~window} \\
\end{tabular}
\caption{A decoded BIMScript program over the input point cloud, wall faces colored by predicted material and openings by role. On this 18-element, three-room validation scene every ground-truth element is recovered (coverage $1.0$, F1@5cm $0.944$), and $16$ of $18$ carry the correct material, at $0.944$ condition accuracy. The one visible attribute error, the \code{tile} wall at the rear predicted \code{wallpaper}, is the frequent-class collapse that \cref{tab:BIMScriptMetrics} quantifies at corpus scale.}
\label{fig:showcase}
\end{figure}

\subsection{Getting appearance to the material tokens}
\label{sec:lifted}
Materials are an \emph{appearance} property, but SceneScript's geometric point encoder discards color. The design question is whether appearance survives the path from pixels to the material token. We route it through a \emph{lifted-feature} encoder: a frozen 2D backbone extracts a dense feature map per keyframe; features are projected to per-3D-point vectors by pose projection, aggregated across observing keyframes, and appended to the point's XYZ as extra input channels to the sparse encoder. The decoder, tokenizer, and grammar are unchanged.

Two implementation details proved essential. \emph{Occlusion filtering:} naive pose projection paints every surface along a ray, and testing projected points against the ground-truth depth removes those wrong-surface features, 46\% of correspondences on one measured scene. \emph{Feature source:} deep semantic features (ResNet-50~\cite{he2016deep} \code{layer4}) are spatially coarse, whereas materials live in color and texture, so we ablate shallower layers (\code{layer2}/\code{layer3}), DINOv2~\cite{oquab2024dinov2} and DINOv3~\cite{simeoni2025dinov3} dense features, raw per-point RGB, and RGB-appended variants (\cref{tab:material}). Because re-lifting the full corpus is expensive, the feature source is chosen once on a small subset and then fixed before the corpus is lifted. 

%The precomputed lifted features also carry per-point observation counts and ray depths, so an observed-mask channel and depth-based occlusion filtering are available without re-lifting the corpus.

% =====================================================================
\section{FastBIMScript: Efficient Decoding}
\label{sec:fast}

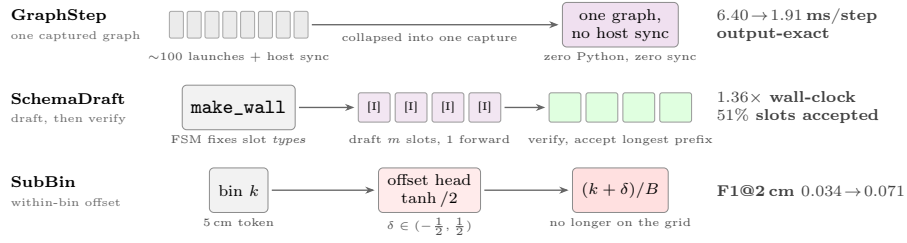
\begin{figure}[tb]
\centering
\resizebox{\linewidth}{!}{%
\begin{tikzpicture}[
  font=\scriptsize,
  blk/.style={draw=black!45, rounded corners=2pt, align=center,
              inner xsep=4pt, inner ysep=1pt, minimum height=7mm},
  krn/.style={draw=black!35, fill=black!8, rounded corners=0.8pt,
              minimum width=2.2mm, minimum height=4.2mm, inner sep=0pt},
  slot/.style={draw=black!40, rounded corners=1pt, minimum width=5mm,
               minimum height=4.6mm, inner sep=0pt, font=\tiny},
  arr/.style={-{Stealth[length=1.5mm]}, semithick, black!60,
            shorten >=1.5pt, shorten <=1.5pt},
  ttlx/.style={font=\scriptsize\bfseries, anchor=west},
  subx/.style={font=\tiny, text=black!60, anchor=west},
  capx/.style={font=\tiny, text=black!70, anchor=north, align=center},
  resx/.style={font=\scriptsize\bfseries, text=black!75, anchor=west, align=left,
               text width=32mm},
]
% shared grid: three stage centers + one result column, used by every row
\def\sA{3.7} \def\sB{6.7} \def\sC{9.7} \def\xR{11.1} \def\cy{-0.52}

% ===================== GraphStep =====================
\def\y{0}
\node[ttlx] at (0,\y+0.14) {GraphStep};
\node[subx] at (0,\y-0.18) {one captured graph};
\foreach \i in {0,...,7} \node[krn] (k\i) at (\sA-0.98+0.28*\i,\y) {};
\node[blk, fill=violet!12] (g3) at (\sC,\y) {one graph,\\no host sync};
\draw[arr] (k7.east) -- (g3.west);
\node[capx] at (\sA,\y+0.2\cy) {${\sim}100$ launches $+$ host sync};
\node[capx] at (\sB,\y+0.5\cy) {collapsed into one capture};
\node[capx] at (\sC,\y+0.2\cy) {zero Python, zero sync};
\node[resx] at (\xR,\y) {$6.40\!\to\!1.91$\,ms/step\\output-exact};

% ===================== SchemaDraft =====================
\def\y{-1.30}
\node[ttlx] at (0,\y+0.14) {SchemaDraft};
\node[subx] at (0,\y-0.18) {draft, then verify};
\node[blk, fill=black!5] (s1) at (\sA,\y) {\code{make\_wall}};
\foreach \i in {0,...,3}
  \node[slot, fill=violet!10] (d\i) at (\sB-0.87+0.58*\i,\y) {[I]};
\foreach \i in {0,...,3}
  \node[slot, fill=green!12]  (v\i) at (\sC-0.87+0.58*\i,\y) {};
\draw[arr] (s1.east) -- (d0.west);
\draw[arr] (d3.east) -- (v0.west);
\node[capx] at (\sA,\y+0.2\cy) {FSM fixes slot \emph{types}};
\node[capx] at (\sB,\y+0.2\cy) {draft $m$ slots, 1 forward};
\node[capx] at (\sC,\y+0.2\cy) {verify, accept longest prefix};
\node[resx] at (\xR,\y) {$1.36\times$ wall-clock\\$51\%$ slots accepted};

% ===================== SubBin =====================
\def\y{-2.60}
\node[ttlx] at (0,\y+0.14) {SubBin};
\node[subx] at (0,\y-0.18) {within-bin offset};
\node[blk, fill=black!5] (b1) at (\sA,\y) {bin $k$};
\node[blk, fill=red!8]   (b2) at (\sB,\y) {offset head\\$\tanh/2$};
\node[blk, fill=red!12]  (b3) at (\sC,\y) {$(k+\delta)/B$};
\draw[arr] (b1) -- (b2);
\draw[arr] (b2) -- (b3);
\node[capx] at (\sA,\y+0.2\cy) {5\,cm token};
\node[capx] at (\sB,\y+0.2\cy) {$\delta\in(-\tfrac12,\tfrac12)$};
\node[capx] at (\sC,\y+0.2\cy) {no longer on the grid};
\node[resx] at (\xR,\y) {F1@2\,cm $0.034\!\to\!0.071$};
\end{tikzpicture}}
\caption{The three FastBIMScript mechanisms. \textit{GraphStep} collapses a
decode step whose cost is kernel-launch and host-synchronization overhead into
a single captured CUDA graph, emitting identical tokens. \textit{SchemaDraft}
exploits the fact that a command token fixes the \emph{types} of all remaining
parameter slots: it drafts them in one forward pass with a learned
\code{[INFILL]} embedding (\code{[I]}) and accepts the longest verified prefix.
\textit{SubBin} keeps the coarse token for structure and adds a bounded
within-bin offset, so a coordinate is no longer pinned to the 5cm grid.}
\label{fig:fastbim}
\end{figure}

\subsection{Where the time actually goes}
Decoding one scene on A100 takes $1.4$--$1.8$\,s against $0.7$--$0.8$\,s for the encoder, so decode is the target. The useful measurement is \emph{per step}, and three observations locate the cost, see \cref{tab:speed}. Per-step time is flat from step 20 to 600, so it does not grow with sequence length. The step is nowhere near FLOP-bound: a $d\,{=}\,512$, 4-layer, single-token step is about 2\,MFLOPs, microseconds of A100 compute. And pre-allocated static buffers are no faster than growing ones, so memory reallocation is not the cost either. What remains is \emph{overhead}: roughly 100 tiny kernel launches per step, plus one host synchronization forced by the type FSM reading a GPU tensor back into Python at every step. Collapsing the step into a single CUDA graph removes exactly that, cutting $5.49$ to $1.91$\,ms/step. The same reasoning explains why KV-caching alone, the obvious first move, buys only $1.2\times$: it removes FLOPs from a loop that was never FLOP-bound.

\subsection{GraphStep: output-exact CUDA-graph decoding}
We compile the entire decode step (embed, self-attention over a static KV buffer, cross-attention over the padded context, feed-forward, final norm, tail projection, argmax, on-device FSM type update, and write-back) into one CUDA graph with zero Python and zero host synchronisation per step; the host syncs only every 16 steps to scan for the \code{stop} token. The FSM grammar is compiled into device lookup tables whose semantics are proven equal to the reference Python FSM up to the first \code{stop} token (CPU fuzz test), and the decoded strings are verified equal to the reference decoder on real scenes. The result is \emph{output-exact}, with identical tokens, and needs no retraining. Per-step cost drops $6.40\!\to\!1.91$\,ms ($3.4\times$); isolating the graph itself from the KV-cache path it builds on, the graph's own contribution is $5.49\!\to\!1.91$\,ms ($2.9\times$). 
%It is restricted to batch 1 with argmax (the production inference path); batched validation falls back to the reference path unchanged.

\begin{table}[tb]
\caption{Decode acceleration, all measured on A100, batch 1, real scenes. \textit{Left}: per-step decode cost; per-step time is flat in sequence length, so the eager$\to$graph gap is overhead, not compute. \textit{Right}: whole-scene wall-clock factors (64 scenes, greedy), smaller than per-step factors because sequence assembly and grammar bookkeeping run outside the graph.}
\label{tab:speed}
\centering
\small
\setlength{\tabcolsep}{3pt}
\setlength{\aboverulesep}{0pt}     
\setlength{\belowrulesep}{0pt}
%\resizebox{\linewidth}{!}{
\begin{tabular}{@{}l|c||@{\qquad}l|c@{}}
Decode step variant & ms/step $\downarrow$ & Whole-scene decode & speedup \\
\midrule
reference loop (recompute) & 6.40 & reference (1.31\,s/scene) & $1.0\times$ \\
KV cache (growing buffers)  & 5.43 & KV cache & $1.31\times$ \\
KV cache (static buffers)   & 5.49 & + GraphStep (graph) & $1.55\times$ \\
\textbf{+ GraphStep (graph)} & \textbf{1.91} & + SchemaDraft ($\tau{=}2$) & $1.36\times$ \\
\end{tabular} %}
\end{table}

\subsection{SchemaDraft: grammar-parallel draft-and-verify}
Generic multi-token prediction~\cite{gloeckle2024better,cai2024medusa}, as used
by Fast SceneScript, must \emph{guess} how many future tokens to emit and gate
them with learned confidence heads. Our grammar removes the guesswork: once a
command token is emitted, the FSM fixes the \emph{types} of every remaining
parameter slot of that element. We therefore draft all $m$ remaining values in
one forward pass, putting a single learned \code{[INFILL]} embedding in place of
each not-yet-known value while position and type embeddings stay intact, then
verify them in one teacher-forced pass and accept the longest prefix matching
the verifier's argmax: structural slots exactly, numeric slots within a
tolerance of $\tau{=}2$ bins, writing back the verifier's token rather than the
draft's. The draft is \emph{schema-complete} rather than learned, so there is no
confidence threshold to tune and no extra projection blocks, one embedding row
against roughly $7.5\%$ added parameters. At $\tau{=}2$ the scheme is
tolerance-verified rather than output-exact; a strict $\tau{=}0$ mode would
recover exactness, and we report $\tau{=}2$ because it is the operating point
we measured. An element of $11$--$13$ sequential steps becomes about 4 forwards
at full acceptance, and this composes with GraphStep.

Acceptance, not drafting, is the binding constraint, because coupled
coordinates such as a wall's far corner given its near corner are genuinely
ambiguous in a one-shot draft. Whole-element masking during the fine-tune and
\emph{intra-element redrafting} at decode time lift acceptance from $38\%$ to
$51\%$; \cref{app:decode} gives the mechanism and the evidence that the
residual gap is coordinate-coupling ambiguity rather than undertraining.
Measured over 64 scenes at $\tau{=}2$, sequential forwards per scene drop
$1.33\times$ and decode wall-clock falls $1.31\!\to\!0.97$\,s/scene
($1.36\times$ over the reference loop, without GraphStep fusion), while the
output agrees with strict greedy decoding on $96.6\%$ of tokens
(\cref{tab:speed}). The fine-tune itself does not cost geometry
(\cref{sec:exp}). The same \code{[INFILL]} capability would serve the
human-in-the-loop local-correction task of Xie
\etal~\cite{xie2025localcorrections} at draft-verify speed.

With decode at $0.85$--$0.97$\,s/scene the fp32 sparse encoder ($0.7$\,s) is
already co-dominant and bounds what further decode work can return; an fp16
encoder attempt is a negative result reported in \cref{app:decode}.

% =====================================================================
\section{Exactly Where: Sub-bin Geometric Refinement}
\label{sec:where}

With $B{=}640$ bins over $[0,32]$\,m the grid spacing is 5cm, so nearest-bin quantization of a coordinate has a worst-case error of $2.5$\,cm before any model error is added, below common BIM tolerances~\cite{usibd2019loa}. The first step is to make this measurable: we add tight F1 thresholds ($0.1$--$5$\,cm) as evaluation overrides, never changing the default thresholds that every legacy number depends on. Because the synthetic ground truth is exact, the measurement is clean. We then test two complementary recoveries; the tokens carry topology and semantics, the refinement carries metric precision.

\noindent\textbf{Training-free snapping} reads the point cloud directly. For each wall we collect points within a tolerance of the wall rectangle, fit the 2D line by PCA, and re-project the endpoints onto the fit; doors, windows, and heights follow the same pattern. It is a post-process and sets the bar the learned head must beat.

\noindent\textbf{SubBin} keeps the coarse token for structure and adds a small MLP head on the decoder's last hidden state that regresses a within-bin offset per numeric slot, bounded to $(-0.5,0.5)$ by a $\tanh/2$ activation. Under our round+edge convention the final coordinate is $(\text{bin}+\text{offset})/B$. The head trains with a Huber loss on the pre-rounding residual over numeric slots only, fine-tuned from an existing checkpoint, and its parameters exist only when enabled, so legacy checkpoints load unchanged. At inference one extra teacher-forced pass emits the offsets, composing for free with the SchemaDraft verify pass. \Cref{sec:exp} reports what this recovers, which is real but small, and \cref{app:decode} gives the full per-type table.

% =====================================================================
\section{Experiments}
\label{sec:exp}

%%---- table 3 ----%%%
\begin{table}[tb]
\caption{Appearance-source ablation for material and condition. All six arms share one 2{,}250-scene training subset, one 20k-step budget, and the same 250 validation scenes, so the comparison isolates the appearance source. Condition macro-F1 is over the three classes every arm's matched population realizes (\code{good}, \code{new}, \code{worn}). Single-seed throughout, so we read the ordering as evidence rather than a settled ranking.}
\label{tab:material}
\centering
\small
\setlength{\tabcolsep}{3pt}
\setlength{\aboverulesep}{0pt}     
\setlength{\belowrulesep}{0pt}     
\resizebox{\linewidth}{!}{
\begin{tabular}{@{}l|l|c|cc|cc@{}}
\multirow{2}{*}{Encoder} & \multirow{2}{*}{Appearance source}
  & \multirow{2}{*}{Coverage $\uparrow$}
  & \multicolumn{2}{c|}{Material} & \multicolumn{2}{c}{Condition} \\
 & & & Accuracy $\uparrow$ & Macro-F1 $\uparrow$ & Accuracy $\uparrow$ & Macro-F1 $\uparrow$ \\
\midrule
point-only & none (geometry)        & 0.269 & 0.326 & 0.259 & 0.715 & 0.322 \\
lifted & raw per-point RGB          & 0.276 & 0.305 & 0.220 & 0.701 & 0.324 \\
lifted & ResNet-50 \code{layer2}    & 0.260 & 0.347 & 0.239 & 0.734 & 0.337 \\
lifted & ResNet-50 \code{layer3}    & 0.259 & 0.341 & 0.256 & 0.729 & 0.348 \\
lifted & DINOv3 ViT-S/16            & 0.253 & \textbf{0.370} & 0.245 & 0.742 & 0.356 \\
lifted & DINOv2 ViT-S/14            & 0.267 & 0.369 & \textbf{0.257} & \textbf{0.769} & \textbf{0.366} \\
\end{tabular}}
\end{table}

\noindent\textbf{Data and protocol.} We use the synthetic indoor corpus of \cref{sec:passport} with the splits of \cref{tab:corpus}, and evaluate on the full $1{,}000$-scene validation and test splits. % unless a subset is stated. 
We report layout F1 under both our dense-threshold protocol and a local, FSS-style protocol (\cref{sec:parity}); material and condition macro-F1 over element pairs matched by the Hungarian algorithm~\cite{kuhn1955hungarian} at a 10\,cm threshold; and decode latency at batch 1 (single L40S for \cref{tab:BIMScriptMetrics},
single A100 for \cref{tab:speed}). All models optimize with AdamW~\cite{loshchilov2019decoupled} at learning rate $10^{-4}$ and effective batch 64, following the original recipe. Every number we report is decoded greedily at a fixed decode batch size of 4. 
%, including the public-checkpoint rows, so all rows of a table share one protocol; we state this because decode batch size measurably shifts these metrics, and the reference implementation decodes one scene at a time.
Material models fine-tune from a layout checkpoint by type-embedding row surgery, which lets the material grammar reuse a trained layout decoder without renumbering the type enumeration; the \textit{SchemaDraft} and \textit{SubBin} fine-tunes descend from the same reproduction checkpoint.

\noindent\textbf{Adding materials does not cost geometry.} The full BIMScript model emits material and condition alongside every geometric parameter, so the first question is what that costs in layout accuracy. On the $1{,}000$-scene test split it scores $0.577$ F1@5cm against $0.571$ for the point-only layout model on the same grid, splits, and protocol (\cref{tab:repro}): the material-aware model is marginally \emph{ahead} of the layout-only baseline, on a total budget of 40k lifted layout steps plus a 40k-step material fine-tune against the baseline's 200k. The speed and precision fine-tunes are likewise inside the reproduction run's own eval-to-eval noise band ($0.56$--$0.62$ F1@5cm). Appearance semantics therefore ride along in the same grammar essentially for free, see \cref{fig:showcase}.
% ($0.585$ val)

\noindent\textbf{Appearance decides which attributes are recoverable.}
\Cref{tab:material} reports two experiments. The six-arm ablation holds the subset, the budget, and the matched-class population fixed across arms, so it is the experiment that supports an appearance-source ranking. There, material \emph{macro}-F1 is nearly flat, since every arm leans on the taxonomy's frequent classes at this budget, but accuracy and \emph{condition} macro-F1 (over the three classes every matched population realizes) separate the sources: DINOv2 leads condition F1 ($0.366$ against point-only's $0.322$), DINOv3 edges it on material accuracy ($0.370$ vs $0.369$) while trailing on condition ($0.356$), and raw per-point RGB buys nothing over geometry alone ($0.324$ against $0.322$). A learned, view-stable feature space, not color itself, is what survives lifting and per-point pooling.

% for this para refer to the new table 4
Scaling the winning arm to the full corpus (\cref{tab:BIMScriptMetrics}) is what makes the attributes usable: the full-corpus model matches $70\%$ of ground-truth elements and predicts their material at $0.626$ accuracy on validation and $0.629$ on test. 

Condition macro-F1 is the weaker of the two attributes and nearly identical across splits ($0.353$ val, $0.352$ test), improving with the longer fine-tune ($0.342$ at half the budget). The average is carried almost entirely by \code{good} ($0.914$): \code{new} reaches only $0.144$ and \code{worn}, despite $485$ matched test instances, is never recovered. Condition is therefore predicted at the level of ``is this the default state or not,'' and the rare tail (\code{damaged}, $0.05$\% of the corpus) is a smaller effect than the failure on \code{worn}.

%, and an earlier checkpoint of this same model differed by $0.05$ between the two splits on this metric alone. 
% We have no full-corpus point-only counterpart trained to the same budget, so this block is a capability measurement, not a second appearance comparison; an earlier attempt at a matched-scale head-to-head compared different class supports and entity populations across arms and did not survive audit, and we do not draw a conclusion from it.

%%---- table 4: full BIMScript grammar, against the public anchor ----%%%
\begin{table}[tb]
\caption{The full BIMScript grammar against its reproduction anchor, on the $1{,}000$-scene \emph{test} split. \emph{Public ckpt} (anchor) emits layout only, so its attribute cells are empty by construction. BIMScript$^{+}$ continues the 40k model for 20k steps with the SchemaDraft and SubBin objectives (\cref{sec:fast}, \cref{sec:where}). Layout is scored over all ground-truth elements, attributes only over elements matched within 10\,cm (\cref{sec:exp}). Latency is batch 1 with the GraphStep decoder on one L40S, the production path; all other metrics come from one greedy decode per scene at batch 4. (${^*}$Scene latency tracks emitted program length: the public checkpoint's batch-1 decode emits ${\sim}3\times$ more tokens per scene than BIMScript on the same scenes, while per token its 4-layer decoder is about $2\times$ faster than our 8-layer one.)}
\label{tab:BIMScriptMetrics}
\centering
\small
\setlength{\tabcolsep}{3pt}
\setlength{\aboverulesep}{0pt}
\setlength{\belowrulesep}{0pt}
\resizebox{\linewidth}{!}{
\begin{tabular}{@{}l|cccc|cc|c|cc|cc|cc@{}}
& \multicolumn{4}{c|}{F1@5cm $\uparrow$} & \multicolumn{2}{c|}{AvgF1 $\uparrow$}
& & \multicolumn{2}{c|}{Material} & \multicolumn{2}{c|}{Condition}
& \multicolumn{2}{c@{}}{Model} \\
Model & wall & door & window & mean & ours & FSS & Cov. $\uparrow$
& Acc. $\uparrow$ & F1 $\uparrow$ & Acc. $\uparrow$ & F1 $\uparrow$ & Par. $\downarrow$ & s/sc.$^{*}$ $\downarrow$ \\
\midrule
Public ckpt & \textbf{0.621} & 0.720 & 0.400 & \textbf{0.581}
  & 0.667 & \textbf{0.931} & -- & -- & -- & -- & --
  & \textbf{25.9M} & 1.47 \\
BIMScript (40k) & 0.509 & \textbf{0.752} & \textbf{0.469} & 0.577
  & \textbf{0.676} & 0.907 & \textbf{0.701} & 0.629 & 0.285 & \textbf{0.842} & \textbf{0.352}
  & 58.4M & 1.14 \\
BIMScript$^{+}$ (60k) & 0.513 & \textbf{0.752} & 0.462 & 0.576
  & 0.669 & 0.897 & 0.699 & \textbf{0.639} & \textbf{0.291} & 0.840
  & 0.350 & 58.5M & \textbf{1.12} \\
\end{tabular}}
\end{table}

\noindent\textbf{Speed.} \Cref{tab:speed} gives the decode results, all measured on the same hardware and scenes: \textit{GraphStep} is output-exact, \textit{SchemaDraft} tolerance-verified, and each is reported against the same reference loop. The two compose in principle but not yet in one binary: a draft covers $m$ slots where $m$ varies by command and shrinks during redrafting, so fusing it into GraphStep's capture needs a family of graphs keyed by $m$ rather than a single-token graph. We report the two separately and do not assume the factors multiply.

\noindent\textbf{Precision.} The offset head does what the mechanism was designed to do: mean F1@2\,cm doubles, $0.034\!\to\!0.071$, and door and window MAE fall by $10.3\%$ and $3.2\%$, while the training-free snapping baseline it is measured against barely moves. Two further measurements then locate where the remaining error actually lives, which is the more useful outcome. Wall MAE stays near $8.7$\,cm, far above the $1.25$\,cm quantization alone predicts, so at this operating point wall error is dominated by something other than grid pitch, most plausibly residual mis-association; we have not isolated it. And F1@0.5\,cm is zero for every variant. Together these say the 5\,cm grid is no longer the binding constraint on precision: sub-centimeter output needs better detection and association, not finer regression. All of this is conditional on an already-correct detection (\cref{tab:precision}, \cref{app:decode}), the scope the head is designed to operate in.

\noindent\textbf{BIMScript for everything.} The speed and precision capabilities above are measured as separate fine-tunes; BIMScript$^{+}$ folds them into the deployed model, continuing the 40k checkpoint for 20k steps with the \code{[INFILL]} and sub-bin objectives active. The combined fine-tune is metrically free at the strict threshold ($0.576$ vs $0.577$ test F1@5cm, one decode apart), slightly ahead on material accuracy ($0.639$ vs $0.629$), and slightly behind on the coarse averages ($0.669$ vs $0.676$ AvgF1, $0.897$ vs $0.907$ local-FSS), at unchanged batch-1 latency ($1.12$ vs $1.14$\,s/scene with GraphStep). 
A single checkpoint therefore serves plain greedy decoding, draft-and-verify, and sub-bin refinement without giving up the metrics of the specialized models (\cref{tab:BIMScriptMetrics,fig:showcase}).

% =====================================================================
\section{BIM Ingestion and LLM-Driven AEC Reasoning}
\label{sec:application}

\begin{figure}[tb]
\centering
\includegraphics[width=0.98\linewidth]{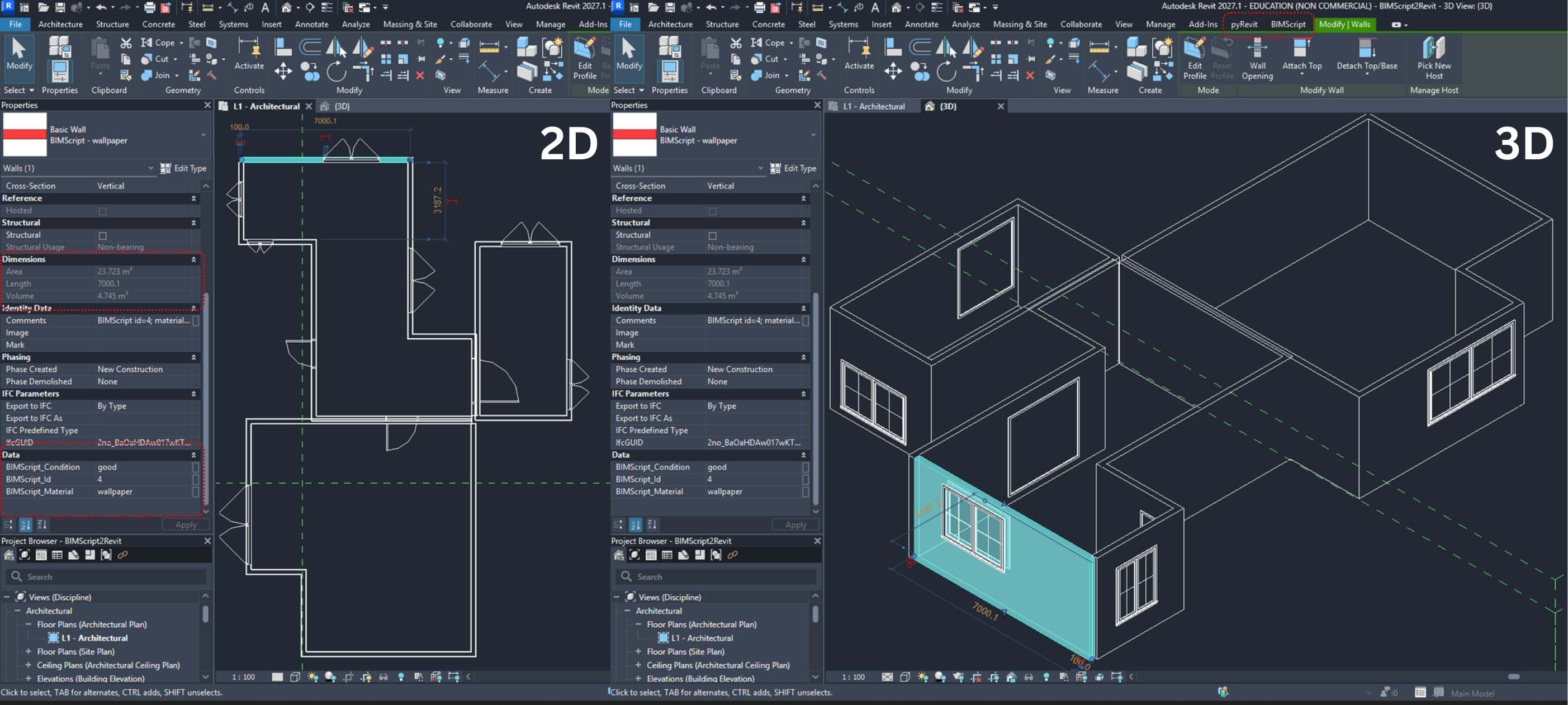}
\caption{A decoded BIMScript program ingested into Autodesk Revit by our pyRevit add-in. The imported plan consists of \emph{native} Revit walls, here an instance of the auto-created type \emph{BIMScript - concrete}, and every element carries its predicted material, condition, and source id as shared parameters.}
\label{fig:revit}
\end{figure}

\noindent\textbf{From program to native BIM.} A BIMScript program is already an object model, not merely a geometric description: each command corresponds one-to-one with a native authoring-tool object, and the material and condition attributes attach to that object's own parameters rather than to a side channel. We implement a deterministic BIMScript-to-Revit adapter as a pyRevit~\cite{PyrevitlabspyRevit} add-in and validate it end-to-end. On the tested Revit and exporter version, importing a decoded program creates \emph{native} Revit walls with one wall type per predicted material, hosted door and window family instances, and the material, condition, and source-element id attached as shared parameters on every element (\cref{fig:revit}). The same document exports to IFC4 through Revit's own exporter, and the exported file's element classes, hosting relations, and BIMScript property set pass independent validation on the evaluated scenes. \Cref{tab:mapping} lists the correspondence between BIMScript commands, Revit API calls, and IFC entities. 
%Parsing, validation, and geometry planning live in a Revit-independent core covered by unit tests, and the add-in is a thin deterministic executor over that plan rather than a general schema translator, so the mapping is auditable rather than implicit.
What this establishes is that native-object ingestion is real, not aspirational; the hard part, getting geometry, material, and condition into one coherent and precise program, is what BIMScript solves.

\noindent\textbf{Language as an interface for reasoning.} Expressing the scene as language makes it a first-class LLM input~\cite{openai2023gpt4}. The material- and condition-annotated program is a compact, machine-readable inventory over which an LLM can reason for sustainability-aware AEC tasks: embodied-carbon and life-cycle estimation (BIM-based LCA~\cite{soust2017critical}), material passports and circularity indicators~\cite{luscuere2017materials,heisel2020calculation,pomponi2017circular}, maintenance triage from the condition attribute (absent from prior structured-scene models), and adaptive-reuse or disassembly decisions. 
% This is a design argument about the representation, not an evaluated result: we have not run a verified LLM query against a decoded program, and leave a checked worked example to future work.

% =====================================================================
\section{Limitations and Future Work}
\label{sec:limitations}
Our supervision is synthetic and its material labels are VLM-distilled pseudo-labels (\cref{sec:passport}), so real-scan generalization and label denoising are open. The rare-class tails are thin in both taxonomies (\code{damaged} at 0.05\% of conditions, \code{glass} at 0.04\% of materials), so macro-F1 depends sharply on which classes a matched population happens to realize. Our reproduction claims are audited against public artifacts rather than a verified replication of SceneScript's Table~2 system (\cref{sec:parity}). GraphStep is batch-1 and argmax by construction, so nucleus sampling and batched decode fall back to the slower path, and SchemaDraft at $\tau{=}2$ is tolerance-verified rather than output-exact. Draft acceptance plateaus at $51\%$ because of coordinate-coupling ambiguity, so the measured $1.36\times$ sits well short of the $3\times$ full-acceptance bound, and the fused GraphStep$\times$SchemaDraft decoder remains future engineering work. Sub-bin refinement is evaluated on exact synthetic ground truth and conditional on correct detection, which flatters it relative to noisy real scans and end-to-end deployment. The Revit and IFC path is validated on programs decoded from synthetic scenes (\cref{sec:application}), not yet against professional as-built modeling on real scans. The grammar itself is the natural thing to grow next: extending it beyond interiors to facades, structural elements such as beams and columns, and mechanical, electrical, and plumbing (MEP) runs would cover what an as-built model actually has to carry, and would need either real scans or synthetic data close enough to stand in for them. Finally, the LLM-driven reasoning we motivate in \cref{sec:application} remains future work.

\begin{table}[tb]
\caption{BIMScript-to-BIM mapping. Both Revit add-in (\cref{fig:revit}), and the IFC exporter through Revit's own IFC4 exporter, whose output we validate independently for element classes, hosting relations, and the BIMScript property set on the evaluated scenes.}
\label{tab:mapping}
\centering
\scriptsize
\setlength{\tabcolsep}{2pt} 
\setlength{\aboverulesep}{0pt}     
\setlength{\belowrulesep}{0pt}
\begin{tabular}{@{}l|l|l@{}}
BIMScript & Revit API target & IFC entity / property set \\
\midrule
\texttt{make\_wall} & \texttt{Wall.Create} (curve, level, height) & \texttt{IfcWall} \\
\texttt{make\_door} & hosted \texttt{FamilyInstance} on host wall & \texttt{IfcDoor} + \texttt{IfcRelFillsElement} \\
\texttt{make\_window} & hosted \texttt{FamilyInstance} on host wall & \texttt{IfcWindow} + \texttt{IfcRelFillsElement} \\
\texttt{material=\dots} & wall/family-type material parameter & \texttt{IfcMaterial}\,/\,\texttt{IfcMaterialLayerSet} \\
\texttt{condition=\dots} & shared parameter (custom) & \begin{tabular}[t]{@{}l@{}} custom Pset (\eg \texttt{Pset\_Condition})
\end{tabular} \\
\end{tabular}
% \vspace{-8mm}
\end{table}

% =====================================================================
\section{Conclusion}
\label{sec:conclusion}
We introduced BIMScript, a structured scene language in which every element carries not only geometry but the material and condition that define a building model, supervised at scale by a VLM-distilled material passport over 1.9M elements. The attributes ride the existing token space and condition on each element's own geometry through self-attention, so they cost nothing to add: the material-aware model matches the layout-only baseline on the same splits. That same design keeps the grammar extensible, since new attributes append without renumbering or decoder changes, and every command still maps one-to-one onto a native Revit object and IFC entity, which we validate end-to-end. Along the way we find that decoding these programs is bound by overhead rather than compute, and that what now limits metric precision is detection rather than the token grid. Representing a building as language brings as-built BIM ingestion within reach, and makes the reconstructed asset a direct input to LLM-driven, sustainability-aware reasoning.

% For the review copy, acknowledgements are omitted to preserve anonymity.
% TODO FINAL: add \section*{Acknowledgements} here for the camera-ready version.

% References start immediately after the main text and do not count toward the
% 14-page limit.
\FloatBarrier
\bibliographystyle{splncs04}
\bibliography{main}

@String(PAMI  = {IEEE Trans. Pattern Anal. Mach. Intell.})

@String(CVPR  = {IEEE Conf. Comput. Vis. Pattern Recog.})

@String(ICCV  = {Int. Conf. Comput. Vis.})

@String(ECCV  = {Eur. Conf. Comput. Vis.})

@String(NeurIPS = {Adv. Neural Inform. Process. Syst.})

@String(ICML  = {Int. Conf. Mach. Learn.})

@String(ICLR  = {Int. Conf. Learn. Represent.})

@String(TMLR  = {Trans. Mach. Learn Res.})

@String(TOG   = {ACM Trans. Graph.})

@String(PAMI  = {IEEE TPAMI})

@String(CVPR  = {CVPR})

@String(ICCV  = {ICCV})

@String(ECCV  = {ECCV})

@String(NeurIPS = {NeurIPS})

@String(ICML  = {ICML})

@String(ICLR  = {ICLR})

@String(TMLR  = {TMLR})

@String(TOG   = {ACM TOG})

@misc{unep_2023,
	author = {United Nations Environment Programme},
	title = {Building Materials and the Climate: Constructing a New Future},
	howpublished ={\url{https://wedocs.unep.org/handle/20.500.11822/43293}},
	year = {2023},
	note = {[Accessed 12-12-2025]},
}

@misc{unhabitat_2023,
	author = {UN-Habitat},
	title = {Future Cities Advisory Outlook 2023: Digital Innovations Empower Urban Net-Zero Carbon Transition},
	howpublished = {\url{https://unhabitat.org/future-cities-advisory-outlook-2023}},
	year = {2023},
	note = {[Accessed 12-11-2025]},
}

@article{BYERS_review_sustrch_2024,
title = {From research to practice: A review on technologies for addressing the information gap for building material reuse in circular construction},
journal = {Sustainable Production and Consumption},
volume = {45},
pages = {177-191},
year = {2024},
issn = {2352-5509},
doi = {https://doi.org/10.1016/j.spc.2023.12.017},
url = {https://www.sciencedirect.com/science/article/pii/S2352550923002920},
author = {Brandon S. Byers and Deepika Raghu and Adama Olumo and Catherine {De Wolf} and Carl Haas}
}

@article{DeWolf_D5_circular_workflow_2024,
title={D5 digital circular workflow: five digital steps towards matchmaking for material reuse in construction}, volume={2}, ISSN={2948-1775}, url={https://www.nature.com/articles/s44296-024-00034-8}, DOI={10.1038/s44296-024-00034-8}, number={1}, journal={npj Materials Sustainability}, author={De Wolf, Catherine and Byers, Brandon S. and Raghu, Deepika and Gordon, Matthew and Schwarzkopf, Vanessa and Triantafyllidis, Eleftherios}, year={2024}, month=nov, pages={36}, language={en} }

@inproceedings{avetisyan2024scenescript,
  author    = {Avetisyan, Armen and Xie, Christopher and Howard-Jenkins, Henry and Yang, Tsun-Yi and Aroudj, Samir and Patra, Suvam and Zhang, Fuyang and Frost, Duncan and Holland, Luke and Orme, Campbell and Engel, Jakob and Miller, Edward and Newcombe, Richard and Balntas, Vasileios},
  title     = {{SceneScript}: Reconstructing Scenes with an Autoregressive Structured Language Model},
  booktitle = ECCV,
  year      = {2024},
  doi       = {10.1007/978-3-031-73337-6_15}
}

@article{yin2025fastscenescript,
  author  = {Yin, Ruihong and Shi, Xuepeng and Bailo, Oleksandr and Manfredi, Marco and Gevers, Theo},
  title   = {Fast {SceneScript}: Fast and Accurate Language-Based {3D} Scene Understanding via Multi-Token Prediction},
  journal = {arXiv preprint arXiv:2512.05597},
  year    = {2025}
}

@article{engel2023aria,
  author  = {Engel, Jakob and Somasundaram, Kiran and Goesele, Michael and Sun, Albert and Gamino, Alexander and Turner, Andrew and others},
  title   = {Project {Aria}: A New Tool for Egocentric Multi-Modal {AI} Research},
  journal = {arXiv preprint arXiv:2308.13561},
  year    = {2023}
}

@inproceedings{avetisyan2020scenecad,
  author    = {Avetisyan, Armen and Khanova, Tatiana and Choy, Christopher and Dash, Denver and Dai, Angela and Nie{\ss}ner, Matthias},
  title     = {{SceneCAD}: Predicting Object Alignments and Layouts in {RGB-D} Scans},
  booktitle = ECCV,
  year      = {2020}
}

@article{tang2010automatic,
  author  = {Tang, Pingbo and Huber, Daniel and Akinci, Burcu and Lipman, Robert and Lytle, Alan},
  title   = {Automatic reconstruction of as-built building information models from laser-scanned point clouds: A review of related techniques},
  journal = {Automation in Construction},
  volume  = {19},
  number  = {7},
  pages   = {829--843},
  year    = {2010}
}

@article{macher2017point,
  author  = {Macher, H{\'e}l{\`e}ne and Landes, Tania and Grussenmeyer, Pierre},
  title   = {From Point Clouds to Building Information Models: {3D} Semi-Automatic Reconstruction of Indoors of Existing Buildings},
  journal = {Applied Sciences},
  volume  = {7},
  number  = {10},
  pages   = {1030},
  year    = {2017}
}

@article{volk2014building,
  author  = {Volk, Rebekka and Stengel, Julian and Schultmann, Frank},
  title   = {Building Information Modeling ({BIM}) for existing buildings --- Literature review and future needs},
  journal = {Automation in Construction},
  volume  = {38},
  pages   = {109--127},
  year    = {2014}
}

@misc{liu2018floornetunifiedframeworkfloorplan,
      title={FloorNet: A Unified Framework for Floorplan Reconstruction from 3D Scans}, 
      author={Chen Liu and Jiaye Wu and Yasutaka Furukawa},
      year={2018},
      eprint={1804.00090},
      archivePrefix={arXiv},
      primaryClass={cs.CV},
      url={https://arxiv.org/abs/1804.00090}, 
}

@misc{stekovic2021montefloorextendingmctsreconstructing,
      title={MonteFloor: Extending MCTS for Reconstructing Accurate Large-Scale Floor Plans}, 
      author={Sinisa Stekovic and Mahdi Rad and Friedrich Fraundorfer and Vincent Lepetit},
      year={2021},
      eprint={2103.11161},
      archivePrefix={arXiv},
      primaryClass={cs.CV},
      url={https://arxiv.org/abs/2103.11161}, 
}

@misc{yue2023connectingdotsfloorplanreconstruction,
      title={Connecting the Dots: Floorplan Reconstruction Using Two-Level Queries}, 
      author={Yuanwen Yue and Theodora Kontogianni and Konrad Schindler and Francis Engelmann},
      year={2023},
      eprint={2211.15658},
      archivePrefix={arXiv},
      primaryClass={cs.CV},
      url={https://arxiv.org/abs/2211.15658}, 
}

@misc{liu2024polyroomroomawaretransformerfloorplan,
      title={PolyRoom: Room-aware Transformer for Floorplan Reconstruction}, 
      author={Yuzhou Liu and Lingjie Zhu and Xiaodong Ma and Hanqiao Ye and Xiang Gao and Xianwei Zheng and Shuhan Shen},
      year={2024},
      eprint={2407.10439},
      archivePrefix={arXiv},
      primaryClass={cs.CV},
      url={https://arxiv.org/abs/2407.10439}, 
}

@misc{xu2024frinetfloorplanreconstructionroomwise,
      title={FRI-Net: Floorplan Reconstruction via Room-wise Implicit Representation}, 
      author={Honghao Xu and Juzhan Xu and Zeyu Huang and Pengfei Xu and Hui Huang and Ruizhen Hu},
      year={2024},
      eprint={2407.10687},
      archivePrefix={arXiv},
      primaryClass={cs.CV},
      url={https://arxiv.org/abs/2407.10687}, 
}

@misc{liu2025cagecontinuityawareedgenetwork,
      title={CAGE: Continuity-Aware edGE Network Unlocks Robust Floorplan Reconstruction}, 
      author={Yiyi Liu and Chunyang Liu and Bohan Wang and Weiqin Jiao and Bojian Wu and Lubin Fan and Yuwei Chen and Fashuai Li and Biao Xiong},
      year={2025},
      eprint={2509.15459},
      archivePrefix={arXiv},
      primaryClass={cs.CV},
      url={https://arxiv.org/abs/2509.15459}, 
}

@misc{armeni2017joint2d3dsemanticdataindoor,
      title={Joint 2D-3D-Semantic Data for Indoor Scene Understanding}, 
      author={Iro Armeni and Sasha Sax and Amir R. Zamir and Silvio Savarese},
      year={2017},
      eprint={1702.01105},
      archivePrefix={arXiv},
      primaryClass={cs.CV},
      url={https://arxiv.org/abs/1702.01105}, 
}

@misc{zheng2020structured3dlargephotorealisticdataset,
      title={Structured3D: A Large Photo-realistic Dataset for Structured 3D Modeling}, 
      author={Jia Zheng and Junfei Zhang and Jing Li and Rui Tang and Shenghua Gao and Zihan Zhou},
      year={2020},
      eprint={1908.00222},
      archivePrefix={arXiv},
      primaryClass={cs.CV},
      url={https://arxiv.org/abs/1908.00222}, 
}

@inproceedings{bieri2025houselayout3d,
  title   = {{HouseLayout3D: A Benchmark and Training-free Baseline for 3D Layout Estimation in the Wild}},
  author  = {Valentin Bieri and Marie-Julie Rakotosaona and Keisuke Tateno and Francis Engelmann and Leonidas Guibas},
  booktitle = {Conference on Neural Information Processing Systems (NeurIPS)},
  year    = {2025}
}

@misc{abouagour2025resplanlargescalevectorgraphdataset,
      title={ResPlan: A Large-Scale Vector-Graph Dataset of 17,000 Residential Floor Plans}, 
      author={Mohamed Abouagour and Eleftherios Garyfallidis},
      year={2025},
      eprint={2508.14006},
      archivePrefix={arXiv},
      primaryClass={cs.CV},
      url={https://arxiv.org/abs/2508.14006}, 
}

@misc{ganon2024wafflemultimodalfloorplanunderstanding,
      title={WAFFLE: Multimodal Floorplan Understanding in the Wild}, 
      author={Keren Ganon and Morris Alper and Rachel Mikulinsky and Hadar Averbuch-Elor},
      year={2024},
      eprint={2412.00955},
      archivePrefix={arXiv},
      primaryClass={cs.CV},
      url={https://arxiv.org/abs/2412.00955}, 
}

@inproceedings{procthor,
  author={Matt Deitke and Eli VanderBilt and Alvaro Herrasti and
          Luca Weihs and Jordi Salvador and Kiana Ehsani and
          Winson Han and Eric Kolve and Ali Farhadi and
          Aniruddha Kembhavi and Roozbeh Mottaghi},
  title={{ProcTHOR: Large-Scale Embodied AI Using Procedural Generation}},
  booktitle={NeurIPS},
  year={2022},
  note={Outstanding Paper Award}
}

@inproceedings{bell2015material,
  author    = {Bell, Sean and Upchurch, Paul and Snavely, Noah and Bala, Kavita},
  title     = {Material Recognition in the Wild with the Materials in Context Database},
  booktitle = CVPR,
  year      = {2015}
}

@inproceedings{upchurch2022dense,
  author    = {Upchurch, Paul and Niu, Ransen},
  title     = {A Dense Material Segmentation Dataset for Indoor and Outdoor Scene Parsing},
  booktitle = ECCV,
  year      = {2022}
}

@article{bell2013opensurfaces,
  author  = {Bell, Sean and Upchurch, Paul and Snavely, Noah and Bala, Kavita},
  title   = {{OpenSurfaces}: A Richly Annotated Catalog of Surface Appearance},
  journal = TOG,
  volume  = {32},
  number  = {4},
  year    = {2013}
}

@article{schwartz2020recognizing,
  author  = {Schwartz, Gabriel and Nishino, Ko},
  title   = {Recognizing Material Properties from Images},
  journal = PAMI,
  volume  = {42},
  number  = {8},
  pages   = {1981--1995},
  year    = {2020}
}

@inproceedings{nash2020polygen,
  author    = {Nash, Charlie and Ganin, Yaroslav and Eslami, S. M. Ali and Battaglia, Peter W.},
  title     = {{PolyGen}: An Autoregressive Generative Model of {3D} Meshes},
  booktitle = ICML,
  year      = {2020}
}

@inproceedings{wu2021deepcad,
  author    = {Wu, Rundi and Xiao, Chang and Zheng, Changxi},
  title     = {{DeepCAD}: A Deep Generative Network for Computer-Aided Design Models},
  booktitle = ICCV,
  year      = {2021}
}

@inproceedings{ganin2021computer,
  author    = {Ganin, Yaroslav and Bartunov, Sergey and Li, Yujia and Keller, Ethan and Saliceti, Stefano},
  title     = {Computer-Aided Design as Language},
  booktitle = NeurIPS,
  year      = {2021}
}

@inproceedings{vaswani2017attention,
  author    = {Vaswani, Ashish and Shazeer, Noam and Parmar, Niki and Uszkoreit, Jakob and Jones, Llion and Gomez, Aidan N. and Kaiser, Lukasz and Polosukhin, Illia},
  title     = {Attention Is All You Need},
  booktitle = NeurIPS,
  year      = {2017}
}

@inproceedings{leviathan2023fast,
  author    = {Leviathan, Yaniv and Kalman, Matan and Matias, Yossi},
  title     = {Fast Inference from Transformers via Speculative Decoding},
  booktitle = ICML,
  year      = {2023}
}

@article{chen2023accelerating,
  author  = {Chen, Charlie and Borgeaud, Sebastian and Irving, Geoffrey and Lespiau, Jean-Baptiste and Sifre, Laurent and Jumper, John},
  title   = {Accelerating Large Language Model Decoding with Speculative Sampling},
  journal = {arXiv preprint arXiv:2302.01318},
  year    = {2023}
}

@inproceedings{cai2024medusa,
  author    = {Cai, Tianle and Li, Yuhong and Geng, Zhengyang and Peng, Hongwu and Lee, Jason D. and Chen, Deming and Dao, Tri},
  title     = {{Medusa}: Simple {LLM} Inference Acceleration Framework with Multiple Decoding Heads},
  booktitle = ICML,
  year      = {2024}
}

@inproceedings{gloeckle2024better,
  author    = {Gloeckle, Fabian and Youbi Idrissi, Badr and Rozi{\`e}re, Baptiste and Lopez-Paz, David and Synnaeve, Gabriel},
  title     = {Better \& Faster Large Language Models via Multi-token Prediction},
  booktitle = ICML,
  year      = {2024}
}

@inproceedings{stern2018blockwise,
  author    = {Stern, Mitchell and Shazeer, Noam and Uszkoreit, Jakob},
  title     = {Blockwise Parallel Decoding for Deep Autoregressive Models},
  booktitle = NeurIPS,
  year      = {2018}
}

@article{shazeer2019fast,
  author  = {Shazeer, Noam},
  title   = {Fast Transformer Decoding: One Write-Head is All You Need},
  journal = {arXiv preprint arXiv:1911.02150},
  year    = {2019}
}

@misc{gray2019cudagraphs,
  author       = {Gray, Alan},
  title        = {Getting Started with {CUDA} Graphs},
  howpublished = {NVIDIA Developer Technical Blog},
  note         = {\url{https://developer.nvidia.com/blog/cuda-graphs/}},
  year         = {2019}
}

@inproceedings{he2016deep,
  author    = {He, Kaiming and Zhang, Xiangyu and Ren, Shaoqing and Sun, Jian},
  title     = {Deep Residual Learning for Image Recognition},
  booktitle = CVPR,
  year      = {2016}
}

@article{oquab2024dinov2,
  author  = {Oquab, Maxime and Darcet, Timoth{\'e}e and Moutakanni, Th{\'e}o and Vo, Huy and Szafraniec, Marc and Khalidov, Vasil and others},
  title   = {{DINOv2}: Learning Robust Visual Features without Supervision},
  journal = TMLR,
  year    = {2024}
}

@article{simeoni2025dinov3,
  author  = {Sim{\'e}oni, Oriane and Vo, Huy V. and Seitzer, Maximilian and Baldassarre, Federico and Oquab, Maxime and Jose, Cijo and others},
  title   = {{DINOv3}},
  journal = {arXiv preprint arXiv:2508.10104},
  year    = {2025}
}

@inproceedings{tang2022torchsparse,
  author    = {Tang, Haotian and Liu, Zhijian and Li, Xiuyu and Lin, Yujun and Han, Song},
  title     = {{TorchSparse}: Efficient Point Cloud Inference Engine},
  booktitle = {Machine Learning and Systems (MLSys)},
  year      = {2022}
}

@inproceedings{choy20194d,
  author    = {Choy, Christopher and Gwak, JunYoung and Savarese, Silvio},
  title     = {{4D} Spatio-Temporal {ConvNets}: {Minkowski} Convolutional Neural Networks},
  booktitle = CVPR,
  year      = {2019}
}

@inproceedings{dai2017scannet,
  author    = {Dai, Angela and Chang, Angel X. and Savva, Manolis and Halber, Maciej and Funkhouser, Thomas and Nie{\ss}ner, Matthias},
  title     = {{ScanNet}: Richly-annotated {3D} Reconstructions of Indoor Scenes},
  booktitle = CVPR,
  year      = {2017}
}

@article{openai2023gpt4,
  author  = {OpenAI},
  title   = {{GPT-4} Technical Report},
  journal = {arXiv preprint arXiv:2303.08774},
  year    = {2023}
}

@article{gemmateam2025gemma3,
  author  = {{Gemma Team}},
  title   = {{Gemma 3} Technical Report},
  journal = {arXiv preprint arXiv:2503.19786},
  year    = {2025}
}

@misc{iso16739,
  author       = {{International Organization for Standardization}},
  title        = {{ISO} 16739-1:2018 --- Industry Foundation Classes ({IFC}) for data sharing in the construction and facility management industries},
  howpublished = {ISO Standard},
  year         = {2018}
}

@misc{usibd2019loa,
  author       = {{U.S. Institute of Building Documentation}},
  title        = {{USIBD} Level of Accuracy ({LOA}) Specification Guide, Version 3.0},
  howpublished = {Specification C220-2019},
  year         = {2019}
}

@misc{Abualdenien_LOD_validation_2020,
  title={Formal analysis and validation of Levels of Geometry (LOG) in building information models},
  author={Jimmy Abualdenien and Andr{\'e} Borrmann},
  year={2020},
  url={https://api.semanticscholar.org/CorpusID:220493703}
}

@article{EleannaReliabilityInBIM,
author = {Eleanna Panagoulia  and Tarek Rakha },
title = {Data Reliability in BIM and Performance Analytics: A Survey of Contemporary AECO Practice},
journal = {Journal of Architectural Engineering},
volume = {29},
number = {2},
pages = {04023006},
year = {2023},
doi = {10.1061/JAEIED.AEENG-1483},
URL = {https://ascelibrary.org/doi/abs/10.1061/JAEIED.AEENG-1483}
}

@article{luscuere2017materials,
  author  = {Luscuere, Lars M.},
  title   = {Materials Passports: Optimising value recovery from materials},
  journal = {Proceedings of the Institution of Civil Engineers --- Waste and Resource Management},
  volume  = {170},
  number  = {1},
  pages   = {25--28},
  year    = {2017}
}

@article{honic2019improving,
  author  = {Honic, Meliha and Kovacic, Iva and Rechberger, Helmut},
  title   = {Improving the recycling potential of buildings through Material Passports ({MP}): An Austrian case study},
  journal = {Journal of Cleaner Production},
  volume  = {217},
  pages   = {787--797},
  year    = {2019}
}

@article{heisel2020calculation,
  author  = {Heisel, Felix and Rau-Oberhuber, Sabine},
  title   = {Calculation and evaluation of circularity indicators for the built environment using the case studies of {UMAR} and {Madaster}},
  journal = {Journal of Cleaner Production},
  volume  = {243},
  pages   = {118482},
  year    = {2020}
}

@article{pomponi2017circular,
  author  = {Pomponi, Francesco and Moncaster, Alice},
  title   = {Circular economy for the built environment: A research framework},
  journal = {Journal of Cleaner Production},
  volume  = {143},
  pages   = {710--718},
  year    = {2017}
}

@article{soust2017critical,
  author  = {Soust-Verdaguer, Bernardette and Llatas, Carmen and Garc{\'i}a-Mart{\'i}nez, Antonio},
  title   = {Critical review of {BIM}-based {LCA} method to buildings},
  journal = {Energy and Buildings},
  volume  = {136},
  pages   = {110--120},
  year    = {2017}
}

@misc{naikade2026xrai,
  author       = {Naikade, Prakash and
                  Frederik, Larsen and
                  Jensen, Mads Brath and
                  Elmholdt, Kasper T. and
                  Møgelmose, Andreas},
  title        = {XRAI for Intelligent and Sustainable AEC: Survey
                   on Integrating Extended Reality and Artificial
                   Intelligence for Human-Centric Built Environments
                  },
  month        = may,
  year         = 2026,
  publisher    = {Zenodo},
  doi          = {10.5281/zenodo.20283895},
  url          = {https://doi.org/10.5281/zenodo.20283895},
}

@inproceedings{loshchilov2019decoupled,
  author    = {Loshchilov, Ilya and Hutter, Frank},
  title     = {Decoupled Weight Decay Regularization},
  booktitle = ICLR,
  year      = {2019}
}

@article{kuhn1955hungarian,
  author  = {Kuhn, Harold W.},
  title   = {The {Hungarian} method for the assignment problem},
  journal = {Naval Research Logistics Quarterly},
  volume  = {2},
  number  = {1--2},
  pages   = {83--97},
  year    = {1955}
}

@article{xie2025localcorrections,
  author    = {Xie, Christopher and Avetisyan, Armen and Howard-Jenkins, Henry and Siddiqui, Yawar and Straub, Julian and Newcombe, Richard and Balntas, Vasileios and Engel, Jakob},
  title     = {Human-in-the-Loop Local Corrections of 3D Scene Layouts via Infilling},
  journal   = {arXiv preprint arXiv:2503.11806},
  year      = {2025}
}

@misc{PyrevitlabspyRevit,
	author = {Pyrevit Labs},
	title = {{G}it{H}ub - pyrevitlabs/py{R}evit: {R}apid {A}pplication {D}evelopment ({R}{A}{D}) {E}nvironment for {A}utodesk {R}evit® --- github.com},
	howpublished = {\url{https://github.com/pyrevitlabs/py{R}evit}},
	year = {},
	note = {[Accessed 26-07-2026]},
}

% ---------------------------------------------------------------
% Supplementary material. Placed after the references so it does not count
% toward the 14-page (15 for camera ready) main-paper limit. For the official submission this may
% instead be moved to a separate supplemental PDF.
\clearpage
\appendix
\section*{Supplementary Material}
\setcounter{section}{0}
\renewcommand{\thesection}{\Alph{section}}
\renewcommand{\theHsection}{appendix.\Alph{section}} 

\section{Material Language and Passport Details}
\label{app:material}

% \noindent\textbf{Taxonomies.} The full material taxonomy, in append-only token order, is \\
% $[$\,\code{aluminum},\allowbreak{}
% \code{brick},\allowbreak{}
% \code{composite},\allowbreak{}
% \code{concrete},\allowbreak{}
% \code{glass},\allowbreak{}
% \code{metal},\allowbreak{}
% \code{painted\_plaster},\allowbreak{}
% \code{pvc},\allowbreak{}
% \code{stone},\allowbreak{}
% \code{tile},\allowbreak{}
% \code{wallpaper},\allowbreak{}
% \code{wood},\allowbreak{}
% \code{wood\_paneling},\allowbreak{}
% \code{unknown}].
% The condition taxonomy is
% [\code{new},\allowbreak{}
% \code{good},\allowbreak{}
% \code{worn},\allowbreak{}
% \code{damaged},\allowbreak{}
% \code{unknown}\,$]$. 
% Canonicalization strips parenthetical suffixes and maps labels to atomic \code{snake\_case}; out-of-vocabulary labels map to \code{unknown} and are counted. 

% Because the enumeration is append-only, the material type tokens extend the layout enumeration without renumbering, so every layout checkpoint stays loadable and material models warm-start from one.

\begin{figure}[htbp]
\centering
\includegraphics[width=0.98\linewidth]{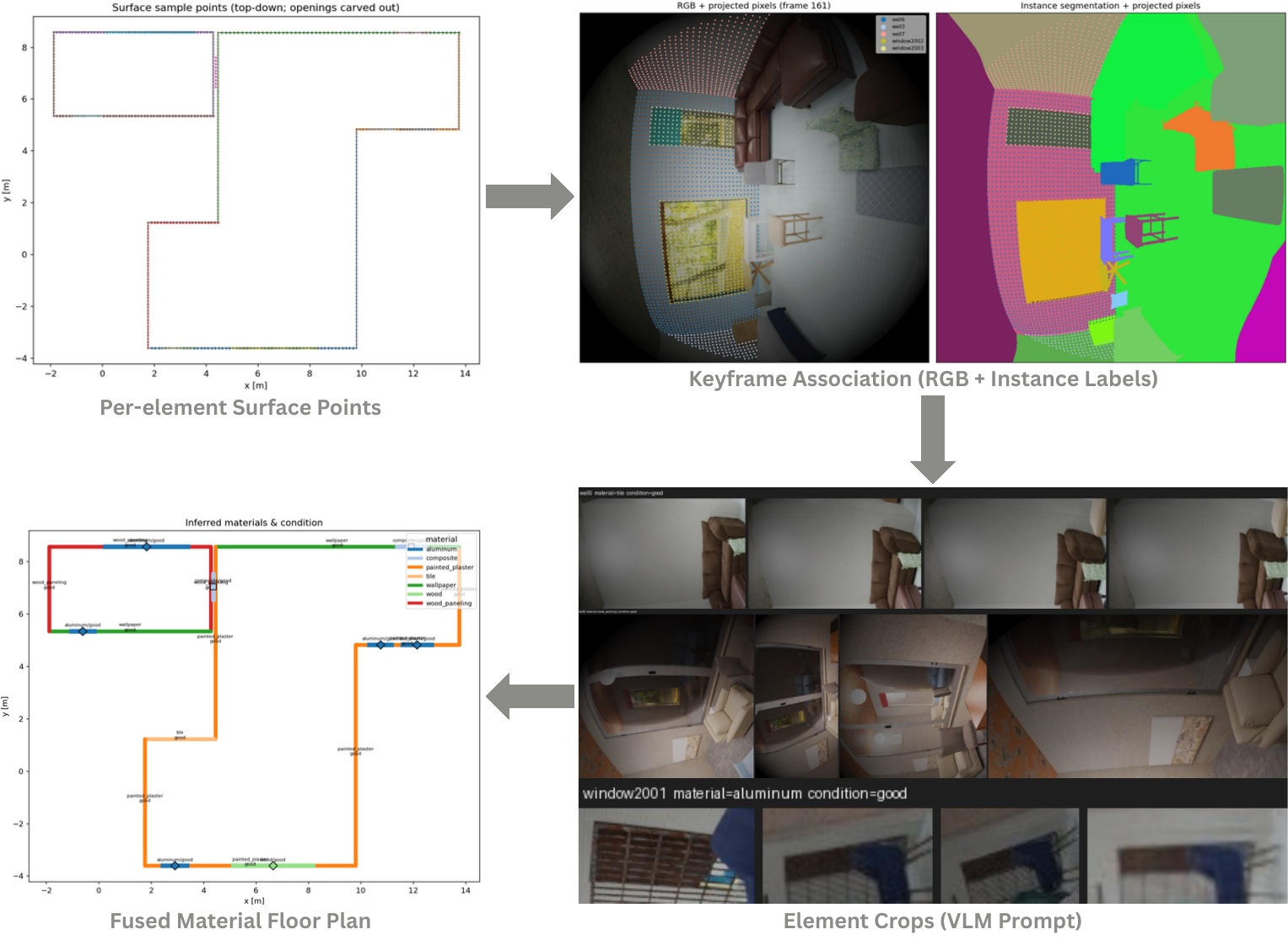}
\caption{Material-passport extraction on one scene. Surface points are associated to elements, the observing keyframes are projected onto each element to cut per-element crops, and the fused per-element label is written back as a material- and condition-annotated floor plan. The crops are what the vision-language model actually sees, which is why occlusion filtering (\cref{sec:lifted}) matters: an unfiltered crop can show the wrong surface entirely.}
\label{fig:materialpipe}
\end{figure}

\noindent\textbf{Extraction.} \Cref{fig:materialpipe} shows the material and condition extraction pipeline on one scene. The label written into the language is the VLM's; 
the dense-segmentation and classical-CV cues are computed and retained in the passport as an independent check rather than merged into it (\cref{tab:corpus}). Labels are canonicalized by stripping parenthetical suffixes and mapping to atomic \code{snake\_case}.
% , which is exactly what the agreement figures in \cref{tab:corpus} measure. 
%Per element we keep the chosen label, all three candidates with their confidences, observation counts, and color/texture statistics. The 
The confidence floor is $0.7$: anything that fails the join to a ground-truth element, is never observed, falls below that floor, or carries an out-of-vocabulary label becomes an explicit \code{unknown} rather than a guessed or a dropped label, and every such case is counted. Composite component break-downs are kept in the passport JSON and never enter the language. 

\begin{table}[htbp]
\caption{The BIMScript dataset and material-passport corpus extracted from ASE dataset. Cross-cue agreement indicates label uncertainty (\cref{sec:passport}).}
\label{tab:corpus}
\centering
\scriptsize
\setlength{\tabcolsep}{4pt}
\setlength{\aboverulesep}{0pt}
\setlength{\belowrulesep}{0pt}
\begin{tabular}{@{}lr|lr|lr@{}}
\multicolumn{2}{@{}l|}{\emph{Corpus}} &
\multicolumn{2}{l|}{\emph{Materials}} &
\multicolumn{2}{l@{}}{\emph{Conditions}} \\
\midrule
Scenes (total)    & 99{,}990      & aluminum          & 20.5\%  & good    & 83.0\% \\
\quad skipped/missing & 10            & wallpaper         & 19.1\%  & new     & 12.9\% \\
\quad train split     & 97{,}990      & painted\_plaster  & 14.0\%  & worn    &  3.7\% \\
\quad val split       & 1{,}000       & wood\_paneling    & 13.1\%  & damaged &  0.05\% \\
\quad test split      & 1{,}000       & wood              & 10.3\%  & unknown &  0.4\% \\
Elements              & 1{,}900{,}908 & composite         &  9.1\%  &         &        \\
Join failures         & 0             & tile              &  6.3\%  &         &        \\
Material classes      & 14            & concrete          &  2.5\%  &         &        \\
Condition classes     & 5             & stone             &  1.7\%  &         &        \\
& & metal  & 1.4\%  & & \\
& & brick &  1.2\%  & & \\
\multicolumn{2}{@{}l|}{\emph{Cross-cue agreement}} & pvc     & 0.3\% & & \\
VLM vs.\ classical CV & ${\sim}25\%$  & glass       & 0.04\% & & \\
VLM vs.\ dense seg.   & ${\sim}34\%$  & unknown           &  0.4\%  &         &    
\end{tabular}
\end{table}

\section{Implementation and Decoding Details}
\label{app:decode}

\noindent\textbf{CUDA-graph.} CUDA-graph decoding uses per-context-length buckets and recaptures only when a scene's context exceeds the current bucket. The FSM device tables encode the parameter-to-next-parameter progression and the command-to-first-parameter mapping; the \code{PART} and \code{STOP} helper tokens override that progression from the sampled value itself, applied branchlessly so the graph never needs a host synchronization to decide a type.

\noindent\textbf{Intra-element redrafting.} One-shot drafts accept only 38\% of
slots, because coupled coordinates such as a wall's far corner given its near corner are genuinely ambiguous before any value is fixed. Two measures recover much of this. First, the infill objective masks \emph{whole elements}, matching draft-time conditioning exactly. Second, after a rejection the remaining slots are redrafted with all accepted values as real context, two forwards per round, so each round converts resolved ambiguity into acceptances and the worst case degrades gracefully toward sequential decoding. Redrafting lifts acceptance to 51\%. Doubling the fine-tune budget changes nothing ($50.8\%$ vs $51.0\%$), which is what indicates the residual gap is coordinate-coupling ambiguity rather than undertraining. Folding the draft and verify passes into GraphStep's graph, and a strict $\tau{=}0$ variant that would recover output-exactness, are future work.

\noindent\textbf{fp16 encoder.} We tried inference-time fp16 autocast on the sparse encoder and measured essentially no gain: the TorchSparse~\cite{tang2022torchsparse} library's custom kernels carry no autocast rules and silently stay fp32, so only the encoder's few dense layers cast. A genuine encoder speed-up needs true fp16 through the sparse kernels, so we deprioritized it.

\noindent\textbf{Sub-bin refinement.} \Cref{tab:precision} reports per-type coordinate MAE and mean F1 at tight thresholds, conditional on an already-correct detection: shared, matched elements only. We do not report an end-to-end precision number over all ground-truth entities, which would additionally penalize missed and spurious detections.

\begin{table}[htbp]
\caption{Sub-bin refinement, conditional on correct detection (64 scenes, shared detections, matched elements). Per-type coordinate MAE (cm) and mean F1 at tight thresholds.}
\label{tab:precision}
\centering
\small
\setlength{\tabcolsep}{3pt} 
\setlength{\aboverulesep}{0pt}     
\setlength{\belowrulesep}{0pt} 
\begin{tabular}{@{}l|ccc|cc@{}}
& \multicolumn{3}{c|}{MAE (cm) $\downarrow$} & \multicolumn{2}{c@{}}{mean F1 $\uparrow$} \\
Method & wall & door & window & F1@1cm & F1@2cm \\
\midrule
no refinement (5cm grid)   & 8.74 & 1.74 & 2.77 & 0.001 & 0.034 \\
+ geometric snapping       & 8.37 & 1.74 & 2.75 & 0.001 & 0.034 \\
+ SubBin offset head       & 8.61 & \textbf{1.56} & \textbf{2.68} & \textbf{0.004} & \textbf{0.071} \\
\end{tabular}
\end{table}

\section{SceneScript Reproduction Audit Details}
\label{app:repro}

\noindent\textbf{Discretization and reconstruction.} The unbiased pairing are floor with center or round-at-discretize with edge reconstruction; either gives per-axis quantization error in $[-2.5,+2.5)$\,cm, and $[-2,+2)$\,cm for height, so a purely arithmetic, model-free round-trip has worst-case per-corner error $\sqrt{2.5^2+2.5^2+2^2}=4.06$\,cm. A controlled A/B on the public checkpoint (100 validation scenes) identifies which convention is consistent with the released weights: edge reconstruction scores $0.570$ F1@5cm while center reconstruction collapses to $0.154$. This identifies a convention consistent with the weights; it is not a direct observation of SceneScript's training recipe.

\noindent\textbf{Translation-frame removals.} The frame's origin is the lowest observed point in the point cloud, so anything the scan did not cover falls at a negative coordinate, outside the range the tokenizer can represent. Ground truth, though, describes the whole floor plan: a room the camera never entered still has walls in it, and those are exactly the ones that land out of range. The tokenizer drops them silently, so the model is trained never to predict them. The cost is therefore a quiet loss of recall rather than a visible error, which is why it went unnoticed until we logged it. Over one live training run this happened 46{,}218 times across 14{,}395 scene-visits, at least $14.5\%$ of the scenes seen, and $92\%$ of what disappeared were walls. Because a run revisits every scene once per epoch, these are removal \emph{events} rather than unique elements; the same wall in the same scene is counted again each time it comes round. Anchoring the frame at the point-cloud minimum \emph{minus} a margin, with a bounded clamp for small overshoots, brings the residual cost down to $0.6\%$ of entities at evaluation time.

\noindent\textbf{Round-trip diagnostic.} This measures the ceiling that the tokenizer imposes on any model. It normalizes, discretizes, and reconstructs the ground truth with the same functions used at train and eval time, then scores the result against the un-round-tripped ground truth (300 validation scenes, 5{,}546
entities, rotation disabled so quantization is the only error source). A model that predicted every bin correctly would score exactly this and no higher. The world frame spans 32\,m, so one bin is 5\,cm at 640 and 1.56\,cm at 2{,}048, and round-to-nearest leaves at most half a bin of per-axis error. At 640 bins the round trip scores $0.994$ mean F1@5cm (per-class $0.987$ wall, $0.999$ door, $0.995$ window). The shortfall from $1.0$ is not rounding, which at $\pm2.5$\,cm per axis stays well inside the 5\,cm threshold, but the $0.6\%$ of entities dropped as out of frame: $1 - 0.006 = 0.994$ recovers the measured value. AvgF1 behaves differently because it averages F1 over thresholds well below one bin width, where quantization does bind: the ceiling is $0.845$ at 640 bins and $0.970$ at 2{,}048. That difference is the headroom the finer grid opens, and the reason its benefit appears at strict thresholds rather than coarse ones. This is a diagnostic rather than a replication of the training-time pipeline, which additionally applies rotation augmentation and
wall/opening host reassignment.

\noindent\textbf{Anchor stability.} The frame origin is the point-cloud minimum less the margin, and the cloud is capped at $500$k points, so for large scenes the origin depends on which points survive subsampling. On the same 300 scenes, $60$ ($20\%$) exceed the cap; across three random draws each, the origin moves by a median of $0.41$\,m in $L_\infty$, a 90th percentile of $1.41$\,m, and at most $2.00$\,m, with $45$ of the $60$ moving more than $5$\,cm. The 90th percentile alone exceeds the $1.0$\,m margin. Cloud and language are translated together, so each sample stays internally consistent and the effect acts as additional translation jitter during training; at evaluation it makes the frame, and hence which entities are clamped or dropped, depend on the draw. The round-trip diagnostic above anchors on the full-cloud minimum, so the $0.6\%$ residual is measured under a stable anchor and is a lower bound on what the live pipeline sees.

\noindent\textbf{Schedule sensitivity.} We trained two from-scratch probes at vocabulary 2{,}048 on the full corpus. Our recipe (LR $10^{-4}$, cosine, 100k steps) stays healthy across classes (window F1@5cm $0.386$); the Fast SceneScript recipe (LR $10^{-3}$, multi-step decay, 92k steps) converges walls and doors ($0.563/0.778$) but collapses windows to $0.058$ and never recovers. The aggressive schedule locks into a wall/door-optimal basin before windows, the most evidence-sparse class, can catch up. We default to cosine.

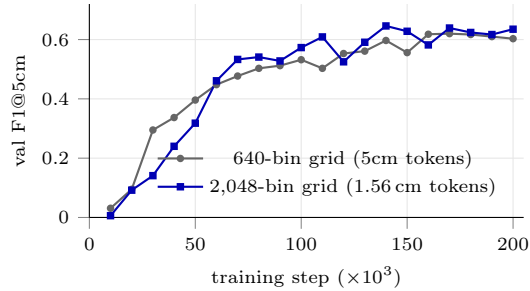
\begin{figure}[tb]
\centering
\begin{tikzpicture}
\begin{axis}[
  width=0.6\linewidth, height=4.4cm,
  xlabel={training step ($\times 10^3$)}, ylabel={val F1@5cm},
  xmin=0, xmax=205, ymin=0, ymax=0.72,
  label style={font=\scriptsize}, tick label style={font=\scriptsize},
  legend style={font=\scriptsize, at={(0.97,0.05)}, anchor=south east,
                draw=none, fill=none},
  axis lines*=left, tick align=outside, grid=major,
  grid style={black!10},
]
\addplot[black!60, thick, mark=*, mark size=1pt] coordinates {
(10,0.031)(20,0.095)(30,0.295)(40,0.337)(50,0.396)(60,0.448)(70,0.477)(80,0.503)(90,0.512)(100,0.532)(110,0.503)(120,0.553)(130,0.561)(140,0.597)(150,0.556)(160,0.618)(170,0.620)(180,0.617)(190,0.610)(200,0.603)
};
\addlegendentry{640-bin grid (5cm tokens)}
\addplot[blue!70!black, thick, mark=square*, mark size=1pt] coordinates {
(10,0.006)(20,0.092)(30,0.141)(40,0.240)(50,0.318)(60,0.461)(70,0.533)(80,0.541)(90,0.528)(100,0.573)(110,0.609)(120,0.525)(130,0.591)(140,0.646)(150,0.628)(160,0.582)(170,0.639)(180,0.624)(190,0.617)(200,0.635)
};
\addlegendentry{2{,}048-bin grid (1.56\,cm tokens)}
\end{axis}
\end{tikzpicture}
\caption{Validation F1@5cm over training for the two language grids (in-run 64-scene slice, identical data, recipe, and step count). The $3.2\times$ finer grid starts slower, since each coordinate needs more optimization, and then overtakes once the model's residual token error shrinks to a few bins, which is exactly where grid pitch decides whether that error passes or fails the 5cm test. The two curves end within the slice's own eval-to-eval noise, consistent with the small margin in \cref{tab:repro}.}
\label{fig:bins}
\end{figure}

\noindent\textbf{Convergence budget.} At an equal 100k budget the finer grid trails at 5cm, since a $3.2\times$ finer vocabulary needs more optimization per coordinate. \Cref{tab:repro} therefore uses a matched 200k budget, where it leads on strict F1 and trails on the coarse average; \cref{fig:bins} traces validation F1@5cm for both grids under identical data, recipe, and step count.

\section{BIMScript-to-BIM Mapping and Additional Results}
\label{app:qual}

% \Cref{tab:mapping} lists the command-to-object correspondence the add-in implements, and \cref{fig:revit} shows a decoded program after ingestion into Revit. 

\begin{figure}[htbp]
\centering
\includegraphics[width=0.94\linewidth]{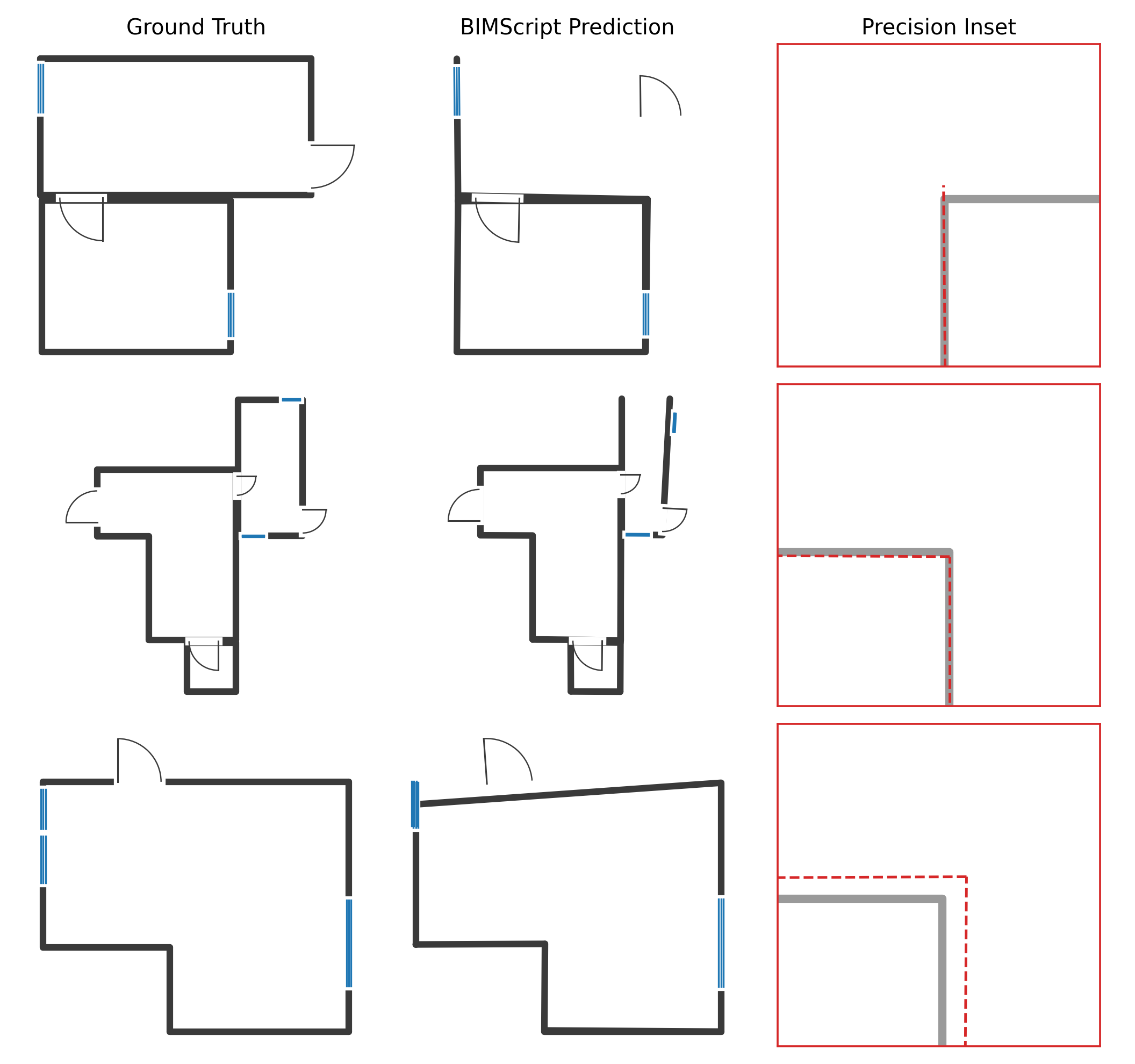}
\caption{Qualitative layout results on the first three validation scenes: ground truth, BIMScript prediction, and a zoomed corner inset (GT solid, prediction dashed); doors are drawn as swing arcs, windows as blue sill glyphs. Rows 1--2 show typical centimeter-level corner agreement; row 3 shows a residual corner offset, and row 1 a missed wall segment, the failure modes behind the strict-F1 numbers in \cref{tab:repro}.}
\label{fig:qual}
\end{figure}

\begin{figure}[tb]
\centering
\renewcommand{\arraystretch}{0.5} 
\setlength{\tabcolsep}{0.pt}
\begin{tabular}{cc}
\footnotesize Ground Truth & \footnotesize BIMScript Prediction \\
\gtpanel{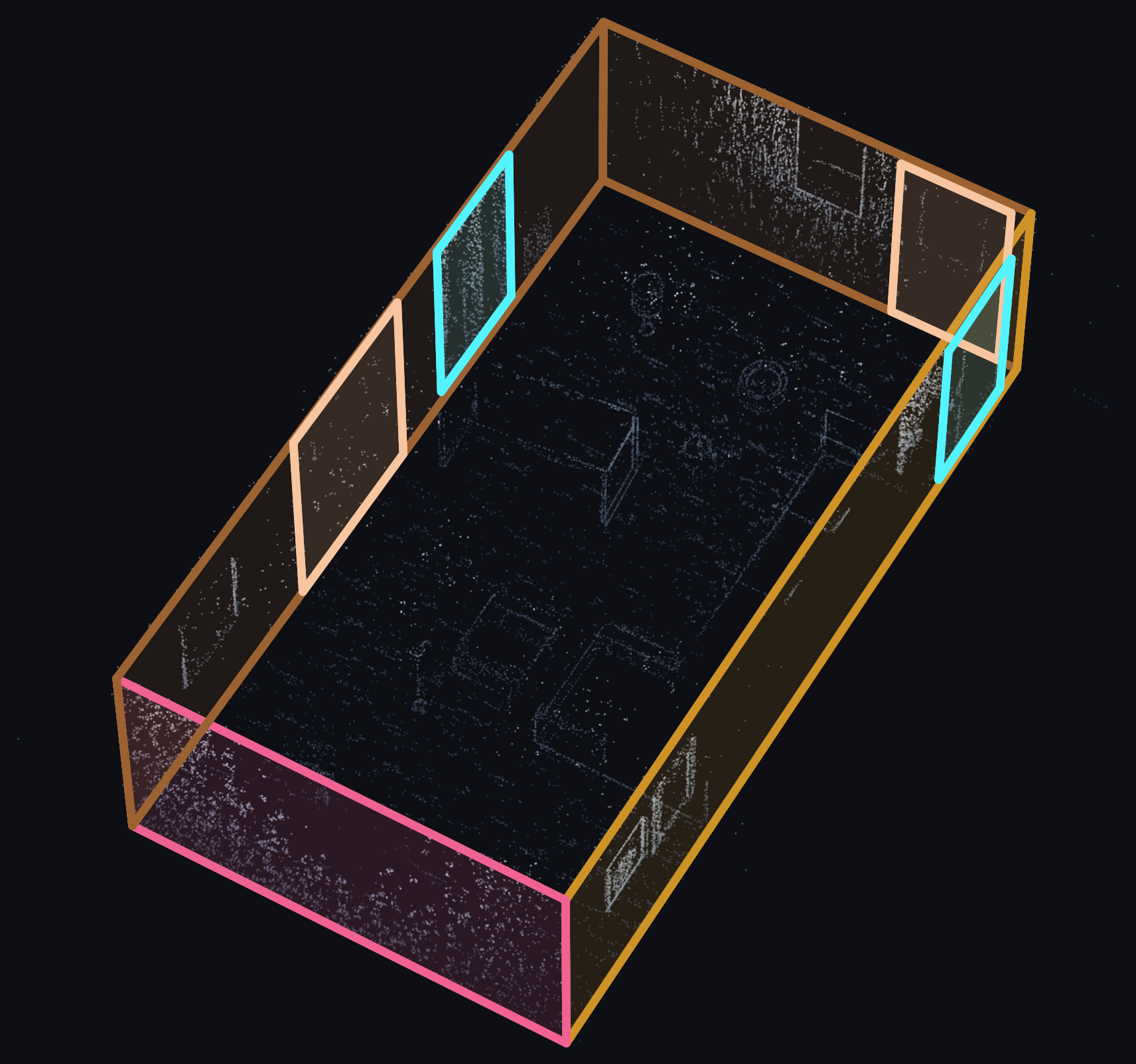}{1} &
\includegraphics[width=0.41\linewidth]{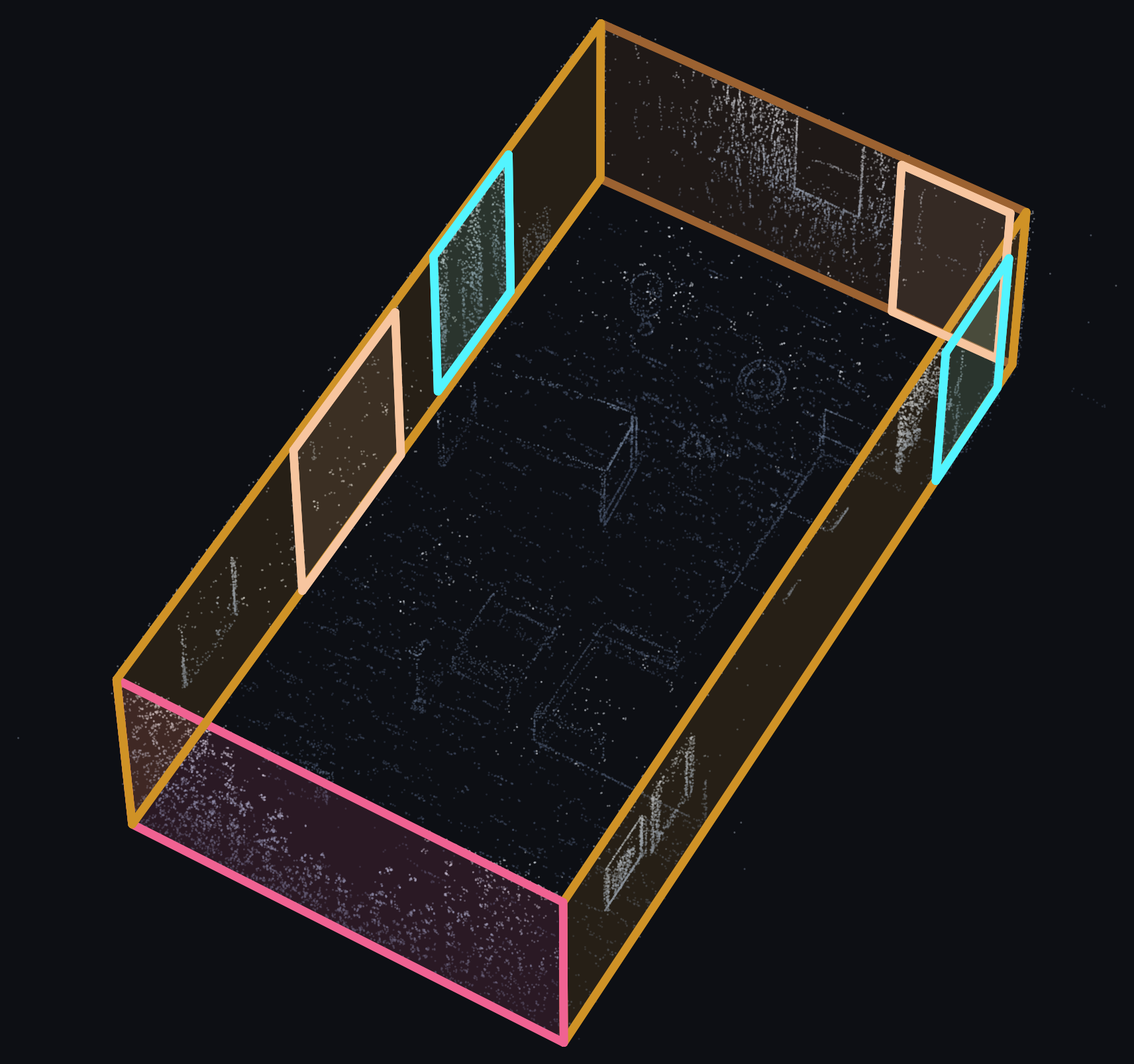} \\
\gtpanel{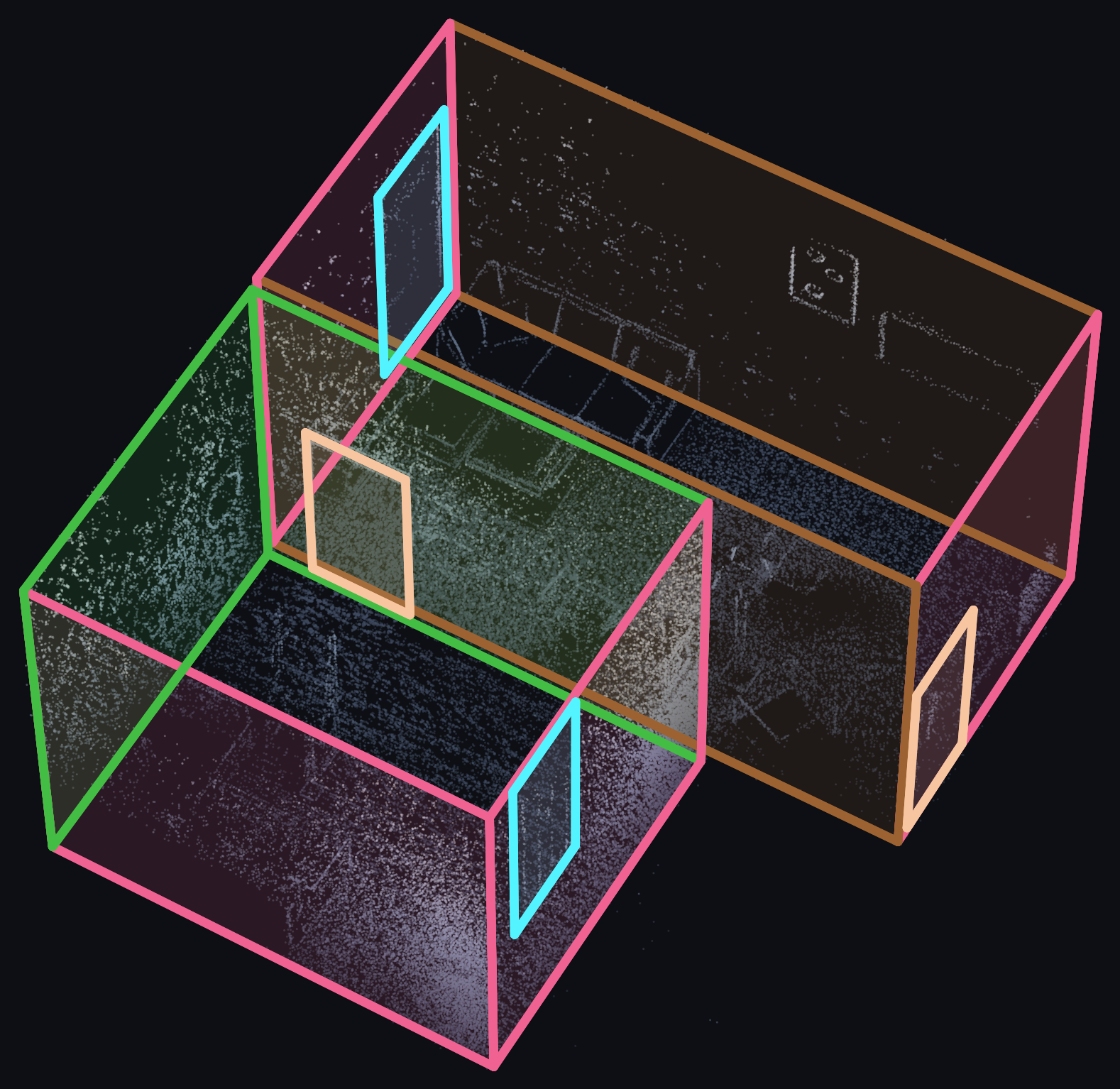}{2} &
\includegraphics[width=0.41\linewidth]{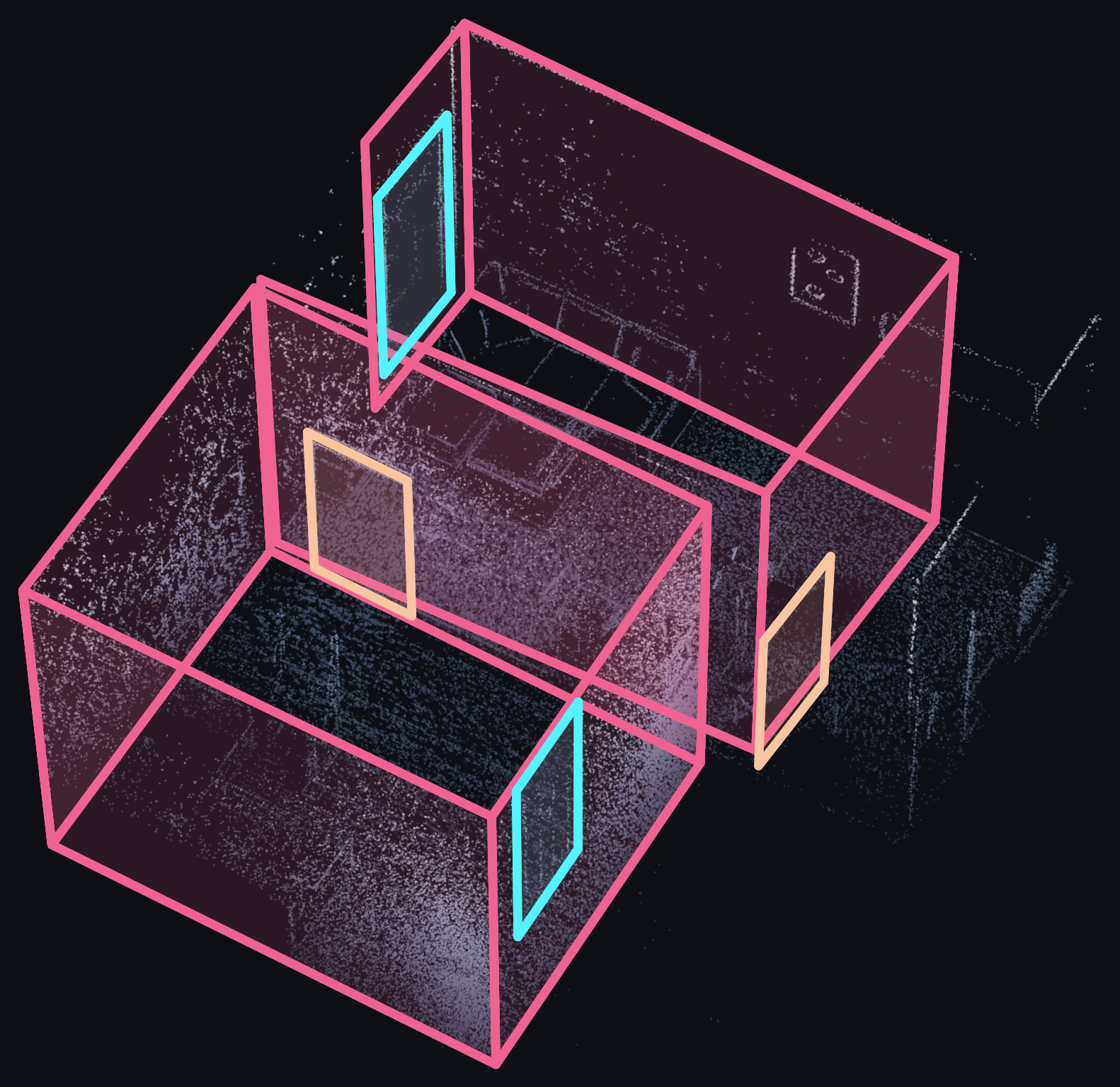} \\
\gtpanel{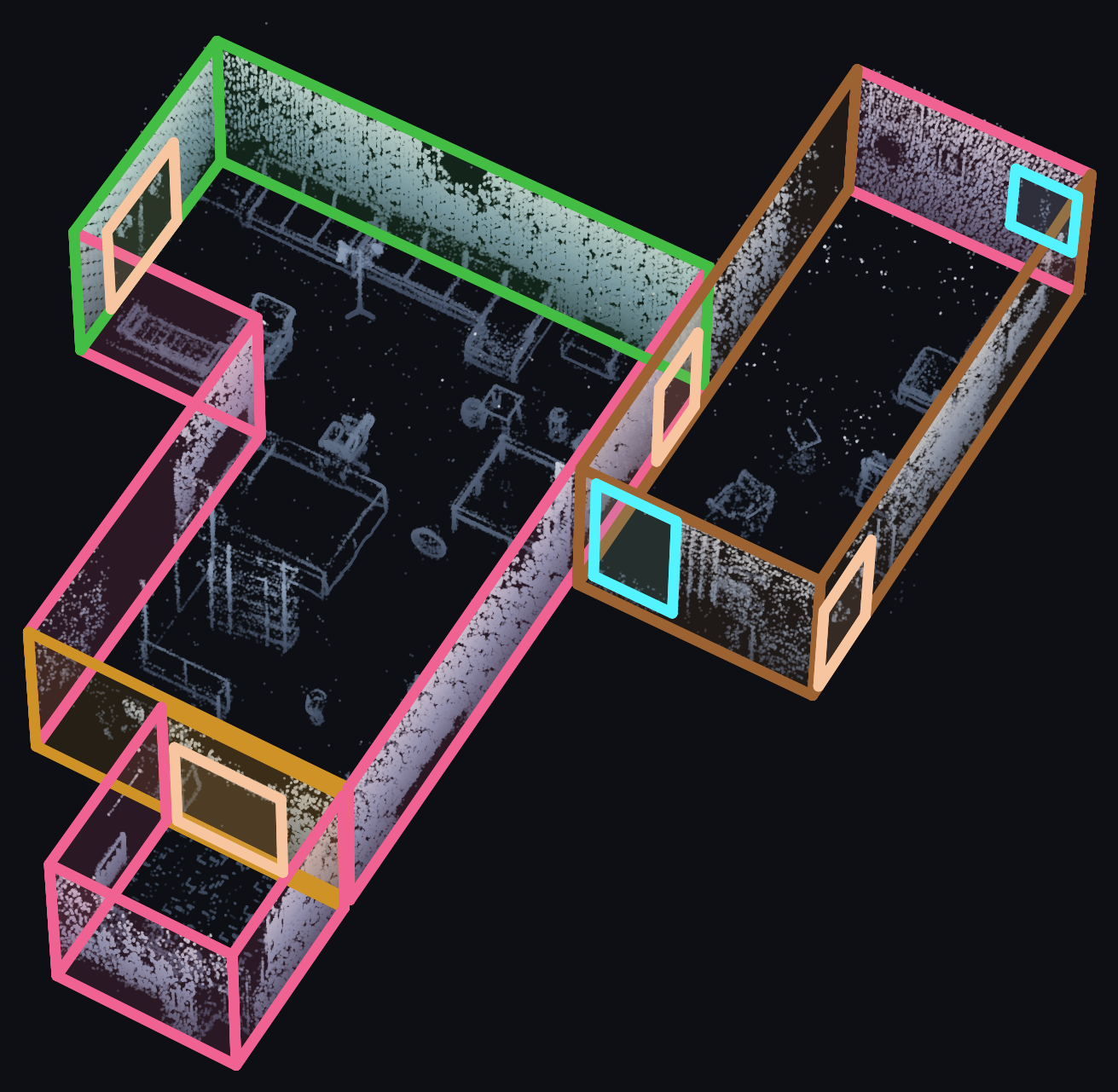}{3} &
\includegraphics[width=0.41\linewidth]{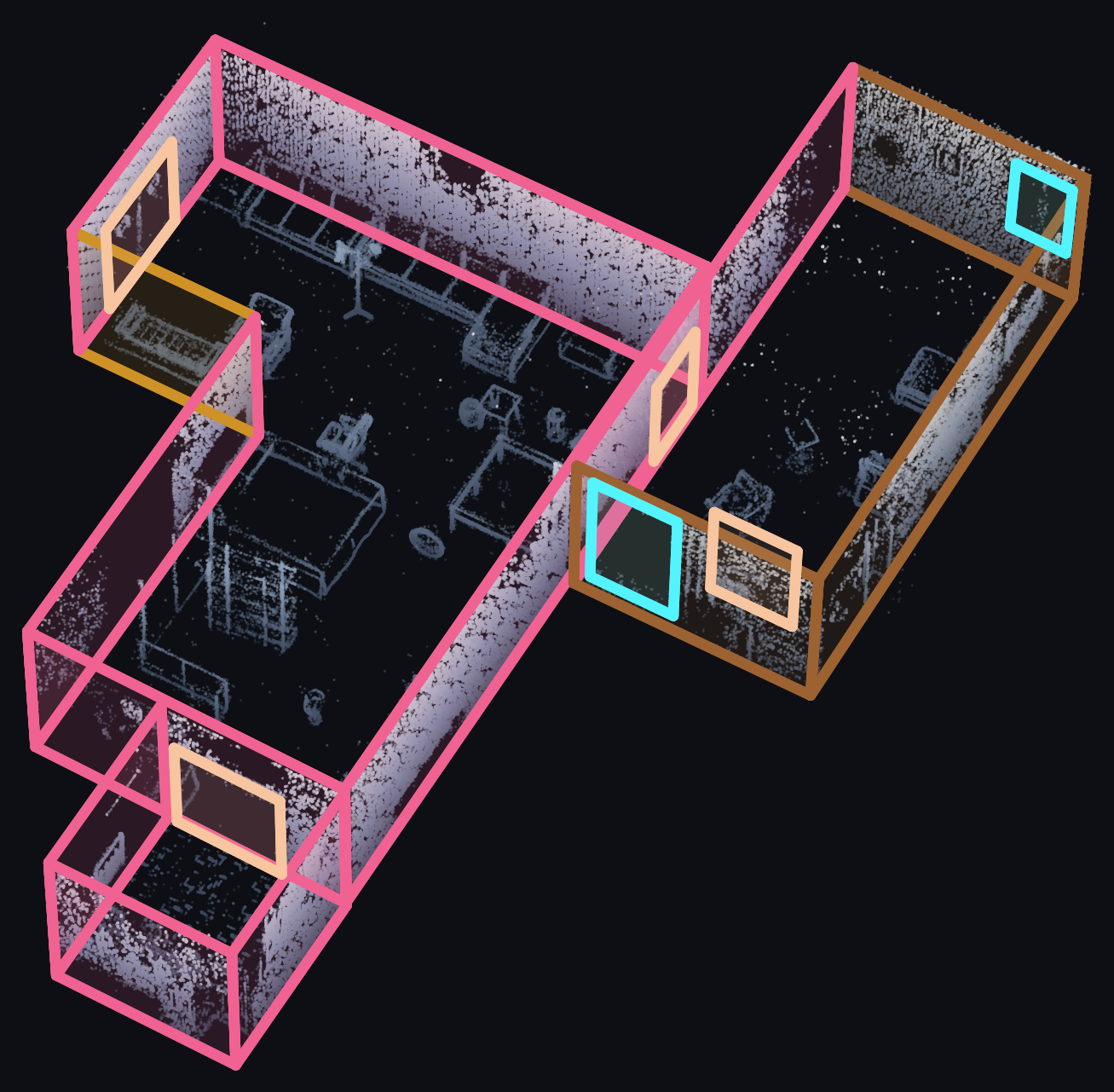} \\
[3pt]
\multicolumn{2}{c}{\scriptsize
  \bsswatch{bsWoodPanR}~wood paneling\,
  \bsswatch{bsWallpaper}~wallpaper\,
  \bsswatch{bsPlaster}~painted plaster\,
  \bsswatch{bsTile}~tile\,
  \bsswatch{bsWood}~wood\,
  \bsswatch{bsDoor}~door\,
  \bsswatch{bsWindow}~window} \\
\end{tabular}
\caption{Decoded programs rendered in 3D over the input point cloud.
% color coding follows \cref{fig:bimscript_teaser}. 
%Rows are validation scenes 714, 22078, and 28395, decoded by the full BIMScript model and not curated. 
Geometry tracks the ground truth closely, and openings are placed on the correct host walls. The materials show both regimes behind the numbers in \cref{tab:material,tab:BIMScriptMetrics}: scene 1 recovers exactly the five materials present in its ground truth, whereas scene 2 collapses tile and wood paneling into the corpus-frequent \code{wallpaper}, which is what a material accuracy of $0.63$ alongside a macro-F1 of $0.29$ looks like element by element. Door and window materials (\code{composite}, \code{aluminum}) are recovered essentially always, since they are near-deterministic given the element class.}
\label{fig:mat3d}
\end{figure}

% 
\Cref{fig:qual} shows uncurated ground-truth/prediction pairs with corner insets for the first three validation scenes. Two failure modes dominate what the strict-F1 numbers measure. Walls bounding \emph{unexplored} floor area are unsupported by points and are either missed or placed with large error, and coordinate error concentrates in $z$, where the range is narrower than in $x$/$y$ so the same bin error is a larger fraction of the extent. \Cref{fig:mat3d} renders decoded programs in 3D with their predicted materials.

\end{document}